\pdfoutput=1

\documentclass[11pt]{article}

\usepackage[final]{acl}

\usepackage{times}
\usepackage{latexsym}

\usepackage[T1]{fontenc}

\usepackage[utf8]{inputenc}

\usepackage{microtype}

\usepackage{inconsolata}

\usepackage{graphicx}

\usepackage{hyperref}
\usepackage{url}
\usepackage{graphicx}
\usepackage{soul}
\usepackage{booktabs}
\usepackage{multirow}

\usepackage{caption}
\usepackage{listings}
\usepackage{xcolor}
\usepackage{upquote}

\usepackage{algorithm}
\usepackage{algpseudocode}

\usepackage{wrapfig}
\usepackage{setspace}

\usepackage{enumitem}

\newcommand{\ie}[0]{\emph{i.e., }}

\newcommand{\eg}[0]{\emph{e.g., }}

\DeclareCaptionFont{white}{\color{white}}
\DeclareCaptionFormat{listing}{%
  \parbox{\linewidth}{\colorbox{gray}{\parbox{\linewidth}{#1#2#3}}\vskip-4pt}}
\usepackage{amsmath,amsfonts,bm}

\def\eqref#1{equation~\ref{#1}}

\def\1{\bm{1}}

\def\vq{{\bm{q}}}

\DeclareMathAlphabet{\mathsfit}{\encodingdefault}{\sfdefault}{m}{sl}
\SetMathAlphabet{\mathsfit}{bold}{\encodingdefault}{\sfdefault}{bx}{n}

\def\gA{{\mathcal{A}}}

\def\gK{{\mathcal{K}}}

\def\gP{{\mathcal{P}}}

\def\gR{{\mathcal{R}}}
\def\gS{{\mathcal{S}}}

\newcommand{\methodname}{{SWAP}}

\title{Deliberate Reasoning in Language Models as Structure-Aware \\Planning with an Accurate World Model}

\author{Siheng Xiong\textsuperscript{1}, Ali Payani\textsuperscript{2}, Yuan Yang\textsuperscript{1*} \& Faramarz Fekri\textsuperscript{1} \\
$^1$Georgia Institute of Technology \quad $^2$Cisco Research\\
\texttt{sxiong45@gatech.edu} \quad 
\texttt{apayani@cisco.com} \quad \texttt{mblackout@hotmail.com}\textsuperscript{*} \\ \texttt{fekri@ece.gatech.edu} 
}

\begin{document}
\maketitle
\begin{abstract}
Enhancing the reasoning capabilities of language models (LMs) remains a key challenge, especially for tasks that require complex, multi-step decision-making where existing Chain-of-Thought (CoT) approaches struggle with consistency and verification.
In this paper, we propose a novel reasoning framework, referred to as \textbf{S}tructure-a\textbf{wa}re \textbf{P}lanning with an Accurate World Model (\methodname{}), that integrates structured knowledge representation with learned planning.
Unlike prior methods that rely purely on natural language reasoning, \methodname{} leverages entailment graphs to encode structured dependencies and enable symbolic verification of intermediate steps. 
To systematically construct and update the graph, \methodname{} employs a policy model to propose candidate expansions and a world model to predict structural updates. 
To improve accuracy, the world model generates multiple alternative updates, and a discriminator re-ranks them based on plausibility. 
To encourage diverse exploration, we introduce Diversity-based Modelling (DM), which samples candidates from the remaining probability mass after removing previously sampled candidates from the original policy distribution.
Additionally, \methodname{} improves the discrimination accuracy through Contrastive Ranking (CR), which directly compares candidates within prompts and incorporates meta-knowledge to improve ranking quality. 
We evaluate \methodname{} across diverse reasoning-intensive benchmarks including math reasoning, logical reasoning, and coding tasks. 
Extensive experiments demonstrate that \methodname{} significantly improves upon the base models and consistently outperforms existing reasoning methods\footnote{Code and data are available at \url{https://github.com/xiongsiheng/SWAP}.}.
\end{abstract}

\section{Introduction}

Achieving human-level problem solving is regarded as the next milestone in Artificial General Intelligence (AGI) \citep{openai2024o1}. 
To enhance the reasoning capabilities of language models (LMs), the Chain-of-Thought (CoT) approach \citep{wei2022chain} is widely adopted due to its scalability and flexibility. 
However, CoT relies solely on natural language reasoning and lacks an effective verification mechanism, leading to inconsistencies and hallucinations in complex reasoning tasks. 
To address this limitation, formal methods such as first-order logic \citep{pan2023logic, cao2025towards} and program-based \citep{chen2022program, ni2024tree} reasoning have been proposed.
Despite their advantages, these methods often lack the expressiveness needed to generalize across diverse reasoning tasks \citep{yang2024can}. 
In this paper, we propose a \textbf{semi-formal reasoning framework} that integrates structured knowledge representation into the reasoning process, enabling symbolic verification of intermediate steps while maintaining flexibility \citep{he2024mintqa, jia2025hetgcot}. 
Our framework represents reasoning as the construction of \textbf{entailment graphs} \citep{dalvi2021explaining}, which explicitly capture how premises lead to intermediate conclusions, facilitating validation of claims. 
Each node in the entailment graph represents a statement, such as evidence, an assumption, or a lemma/rule, while each (hyper)edge represents an entailment relation, mapping a set of premises to a conclusion.
For example, the statements "\textit{All people who regularly drink coffee are dependent on caffeine}" and "\textit{Rina is a student who regularly drinks coffee}" together entail the conclusion "\textit{Rina is dependent on caffeine}".

\begin{figure*}[!t]
    \centering
    \includegraphics[width=0.9\linewidth]{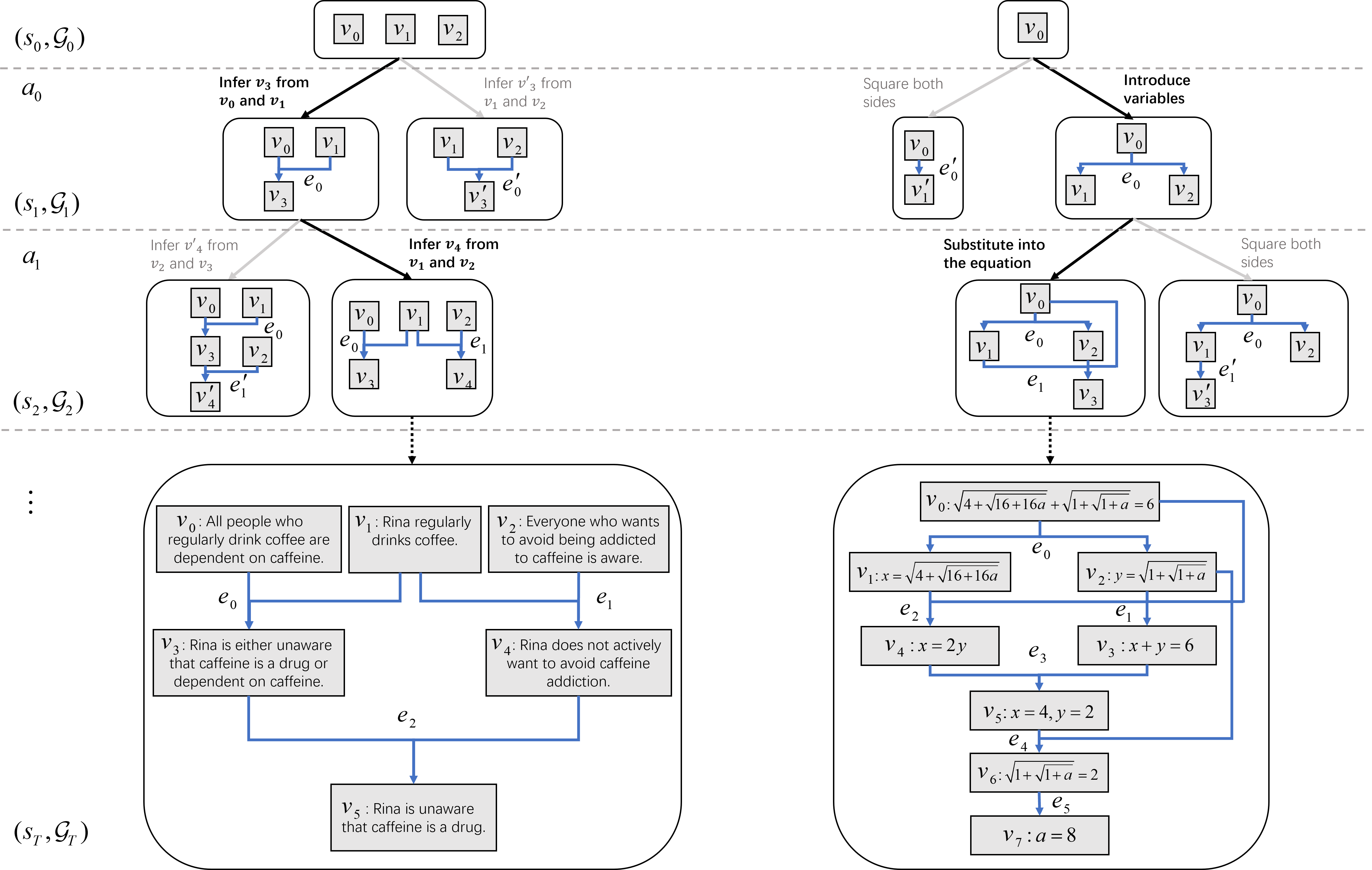}
    \caption{\methodname{} performs multi-step reasoning via structure-aware planning in FOLIO (left) and MATH500 (right). At each step, given the current graph, the policy model proposes an action, while the world model predicts the updated graph as the next state.}
    \label{fig:framework}
\vspace{-10pt}
\end{figure*}

Similar to multi-step reasoning, entailment graph construction can be framed as a sequential decision-making process.
To systematically construct and update the graph, we introduce a planning framework, \textbf{Structure-aware Planning with an Accurate World Model} (\methodname{}), which consists of a policy model and a world model.
In our formulation, the world state corresponds to the graph structure.
Given the current state, the policy model proposes candidate expansions, while the world model predicts state updates.
The accuracy of the LM-based world model is crucial, as it directly impacts the effectiveness of the policy model. 
Existing methods \citep{hao2023reasoning} implement the world model by prompting the same LM with in-context demonstrations, but this approach struggles with complex tasks.
To address this limitation, we enhance the world model by sampling multiple alternative updates and using a discriminator to re-rank them based on plausibility.
Beyond this, we identify two fundamental bottlenecks in planning-based reasoning: \textbf{generation diversity} and \textbf{discrimination accuracy}.  
To encourage diverse exploration, we introduce \textbf{Diversity-based Modelling} (DM), which samples candidates from the remaining probability mass after removing previously sampled candidates from the original policy distribution.
Additionally, \methodname{} improves discrimination accuracy through \textbf{Contrastive Ranking} (CR), which directly compares candidates within prompts and incorporates meta-knowledge to improve ranking quality.
We evaluate \methodname{} across a range of reasoning-intensive benchmarks, including math reasoning, logical reasoning, and coding tasks.
Extensive experiments demonstrate that \methodname{} not only significantly improves upon base models but also consistently outperforms existing reasoning methods.

Specifically, our main contributions include:
\begin{itemize}
\item We introduce \textbf{structure-aware planning}, integrating entailment graphs into multi-step reasoning tasks. These graphs explicitly capture how premises lead to intermediate conclusions, enabling symbolic verification and improving coherence in the reasoning process.
\item We propose \methodname{}, a framework that enhances the policy model with an \textbf{accurate world model}. Additionally, we address two fundamental bottlenecks in planning-based reasoning: generation diversity and discrimination accuracy through Diversity-based Modeling (DM) and Contrastive Ranking (CR), respectively.
\item Extensive experiments on diverse benchmarks, including math reasoning, logical reasoning, and coding tasks, demonstrate that \methodname{} is a generalizable framework that consistently outperforms recent reasoning methods for LMs.
\end{itemize}

\section{Preliminaries}
\subsection{Task formulation}

We formulate the planning task as a Markov Decision Process (MDP) represented as $(\gS, \gA, \gP, \gR)$ where:

\begin{itemize}[noitemsep, topsep=1pt]
\item {\bf State} $s_t \in \gS$: Represents the current graph, capturing all known or inferred information along with entailment relations.
\item {\bf Action} $a_t \in \gA$: Corresponds to an expansion step, where new information is derived or inferred from the current state, leading to a state transition.
\item {\bf Transition probability} $\gP(s_{t+1} | s_{t}, a_{t})$: Defines the probability of transitioning to \( s_{t+1} \) after taking \( a_{t} \) in \( s_{t} \). 
\item {\bf Reward} $\gR(s_t, a_t)$: Measures the quality of \( a_t \) given \( s_t \). 
While reward functions are typically used to select the best action, our framework instead directly compares different actions using a discriminator.
\end{itemize}

This MDP formulation establishes a foundation for applying planning-based methods to enhance the sequential decision-making capabilities of LMs. By iteratively updating their parameters, the models progressively learn an optimal policy, thereby improving their reasoning performance.

\vspace{-2pt}
\subsection{Structured reasoning as entailment graph construction}
\vspace{-2pt}

The key innovation that distinguishes our approach from related work is conceptualizing the multi-step reasoning process as \textbf{entailment graph construction}, which outlines how the premises in $s_0$ lead to intermediate conclusions, ultimately validating the final answer in $s_T$.
Formally, let $\mathcal{G} = (\mathcal{V},\mathcal{E})$ represent the structure, where \(\mathcal{V}\) is the set of nodes, with each node \(v \in \mathcal{V}\) representing a statement, \eg evidence, an assumption, or a lemma/rule; \(\mathcal{E}\) is the set of directed (hyper)edges, where each (hyper)edge \(e = (\mathcal{V}_\text{src}, \mathcal{V}_\text{tgt}) \in \mathcal{E}\) represents an entailment relation from a source node set \( \mathcal{V}_\text{src} \subseteq \mathcal{V} \) (the premises) to a target node set \( \mathcal{V}_\text{tgt} \subseteq \mathcal{V} \) (the conclusions). 

Given $s_0$, the world model $\gP_{\text{wm}}$ first builds the initial graph $\mathcal{G}_0$ by extracting key statements and their relations.
During the reasoning, $\gP_{\text{wm}}$ incrementally grows the graph by adding new nodes and edges, ultimately forming $\mathcal{G}_T$, which includes the final answer.
Additionally, symbolic verification is introduced to ensure the graph quality.
For simplicity, let us denote the state (in natural language) with structural information as $(s, \mathcal{G})$. 
Incorporating this structure provides two main benefits: 1) the policy model can make more informed decisions using the structural information; and 2) the world model can predict more accurate next state.

\vspace{-5pt}
\section{Structure-aware Planning with an Accurate World Model}
\label{sec:method}
\vspace{-2pt}

\subsection{Framework}

{\small
\begin{figure}[!t]
\vspace{-10pt}
\begin{algorithm}[H]
\caption{SWAP($Q, \gP_{\pi}, \gP_{\text{wm}}, \gP_{\text{d}},
\gP_{\text{c}}, N, T, B$)}
\label{alg:swap}
\begin{algorithmic}
\Require Reasoning question $Q$, policy $\gP_{\pi}$, world model $\gP_{\text{wm}}$, discriminator $\gP_{\text{d}}$, controller $\gP_{\text{c}}$, generation number limit $N_{H}$, $N_{a}$ and $N_{s}$, step limit $T$, breadth limit $B$.

\State $\mathcal{D} \gets \{ \}$
\State $G, s_0, \mathcal{G}_0 \gets \texttt{generate}(\gP_{\text{wm}}, Q, 1)$
\State $\mathcal{C} \gets \{ (G, H, s_0, \mathcal{G}_0)\mid H \in \texttt{generate}(\gP_{\pi}, (G,$
\State \hspace{2.5em} $ s_0, \mathcal{G}_0), N_H)\}$

\State $\mathcal{C} \gets \texttt{select}(\gP_{\text{d}}, \{\small{\texttt{simulate}(c, T)} \mid c \in \mathcal{C} \}, B)$

\State $b \gets 0$

\For{$t = 1, \cdots, T$}
\If {$b = B$} \textbf{break}
\EndIf

\State $\mathcal{C} \gets \{ (G, H, s, \mathcal{G}, a) \mid (G, H, s, \mathcal{G}) \in \mathcal{C}, $
\State \hspace{2em} $a \in \texttt{generate}(\gP_{\pi}, (G, H, s, \mathcal{G}), N_a)\}$ 

\State $\mathcal{C}\gets\texttt{select}(\gP_{\text{d}}, \{\small{\texttt{simulate}(c, t) | c \in \mathcal{C}\}}, B)$

\State $\texttt{StatePred}(\mathcal{C}, \mathcal{D}, \gP_{\text{wm}}, \gP_{\text{d}}, \gP_{\text{c}}, N_s, b)$

\EndFor

\State $A^{*} \gets \texttt{select}(\gP_{\text{d}}, \mathcal{D}, 1)$

\State \Return $A^{*}$

\end{algorithmic}
\end{algorithm}
\vspace{-30pt}
\end{figure}

\begin{figure}[!t]
\begin{algorithm}[H]
\setstretch{1.1}
\caption{StatePred ($\mathcal{C}, \mathcal{D}, \gP_{\text{wm}}, \gP_{\text{d}}, \gP_{\text{c}}, N_s, b$)}
\label{alg:statepredict}
\begin{algorithmic}
\Require Context pool $\mathcal{C}$, completed pool $\mathcal{D}$, world model $\gP_{\text{wm}}$, discriminator $\gP_{\text{d}}$, controller $\gP_{\text{c}}$, generation number limit $N_s$, completed reasoning count $b$.

\State \textbf{parallel for} $i = 1, \cdots, |\mathcal{C}|$ \textbf{do} 
    \State \hspace{1em} $(G, H, s, \mathcal{G}, a) = \mathcal{C}_i$
    \State \hspace{1em} $\{(s^{\prime}_j, \mathcal{G}^{\prime}_j)\} \gets \texttt{generate}(\gP_{\text{wm}}, (s, \mathcal{G}, a), N_s)$
    \State \hspace{1em} $ \{(s^{\prime}_{j^{\prime}}, \mathcal{G}^{\prime}_{j^{\prime}})\} \gets \texttt{symCheck}( \{(s^{\prime}_j, \mathcal{G}^{\prime}_j)\})$
    \State \hspace{1em} $s^{\prime}, \mathcal{G}^{\prime} \gets \texttt{select} (\gP_{\text{d}}, \{(s, \mathcal{G}, a, s^{\prime}_{j^{\prime}}, \mathcal{G}^{\prime}_{j^{\prime}})\}, 1)$
    \State \hspace{1em} $A \gets \texttt{extractAns}(\gP_{\text{c}}, (G, s^{\prime}, \mathcal{G}^{\prime}))$
    \State \hspace{1em} \textbf{if} {$A \neq None$} \textbf{then} 
        \State \hspace{2em} $\mathcal{D}.\text{add}((G, s^{\prime}, \mathcal{G}^{\prime}, A))$ \Comment{collect the state}
        \State \hspace{2em} $\mathcal{C}.\text{pop}(i)$ \Comment{remove context $\mathcal{C}_i$}
        \State \hspace{2em} $b \gets b + 1$
\State \hspace{1em} \textbf{else}
        \State \hspace{2em} $\mathcal{C}_i \gets (G, H, s^{\prime}, \mathcal{G}^{\prime})$ \Comment{update context $\mathcal{C}_i$}
\State \hspace{1em} \textbf{end if}
\State \textbf{end for}

\end{algorithmic}
\end{algorithm}
\vspace{-20pt}
\end{figure}
}

In this section, we present our framework (\methodname{}) that enables LMs to systematically construct and utilize an entailment graph for solving a wide range of reasoning tasks.
It integrates \textbf{symbolic structure} into each reasoning state and relies on multiple components to \textbf{iteratively explore, evaluate, and refine reasoning paths} toward correct final answers.
We use $\gP_{\pi}$ to denote the policy, $\gP_{\text{wm}}$ to denote the world model, $\gP_{\text{d}}$ to denote the discriminator, and $\gP_{\text{c}}$ to denote the controller, all based on pre-trained LMs. 
We consider $Q, A, G, H, s, \mathcal{G}, a$ as language sequences, \ie $Q=(Q[1], \cdots, Q[L])$ where each $Q[l]$ is a token, so that $\gP(Q) = \prod_{l=1}^{L} \gP(Q[l] | Q{[1..l-1]})$. 
We use $(s, \mathcal{G})$ to denote the state (in natural language) with structural information, 
and $c = (G, H, s, \mathcal{G})$ to denote the context of goal $G$, plan $H$ and $(s, \mathcal{G})$.

For notational convenience, we define the generation process as $\texttt{generate}(\texttt{model}, \texttt{input}, N)$ where $N$ is the sampling number, the symbolic verification process as $\texttt{symCheck}(\texttt{input})$, and the selection process as $\texttt{select}(\texttt{model}, \texttt{input}, B)$ where $B$ is the number of preserved candidates.
To search potential plans and actions, we simulate the future situations of the current state using the world model.
Specifically, we use $\texttt{simulate}(c, t)$ to denote the simulation starting from $(s, \mathcal{G})$ up to step $t$ given the goal $G$ and plan $H$ from the context $c$.
Following state prediction, we attempt to extract final answers from the resulting state, denoted as $\texttt{extractAns}(\texttt{model}, \texttt{state})$.

Algorithm \ref{alg:swap} outlines the \textbf{workflow}. 
Given a reasoning question \( Q \), the world model $\gP_{\text{wm}}(G,s_0, \mathcal{G}_0|Q)$ first extracts the goal \( G \) and the initial state \( (s_0, \mathcal{G}_0) \).
The policy then proposes a set of plans $H$ by sampling \( N_H \) times from $\gP_{\pi}(H|G, s_0, \mathcal{G}_0)$.
The top \( B \) candidate plans are selected by the discriminator \( \gP_{\text{d}} \) based on the simulation results $(s_T, \mathcal{G}_T)$ under each plan.
Given the goal $G$, selected plan $H$ and current state $(s_{t-1}, \mathcal{G}_{t-1})$, 
expansion at step $t$ begins with the policy sampling \( N_a \) times from $\gP_{\pi}(a_{t-1} | G, H, {s}_{t-1}, \mathcal{G}_{t-1})$ as the next action pool.
The discriminator \( \gP_{\text{d}} \) then evaluates and selects the top \( B \) candidate contexts ($G$, $H$, $s_{t-1}$, $\mathcal{G}_{t-1}$, $a_{t-1}$) based on simulated states $(s_t, \mathcal{G}_t)$.

The \textbf{accurate state prediction} (Algorithm \ref{alg:statepredict}) is then performed in parallel for each selected context ($G$, $H$, $s_{t-1}$, $\mathcal{G}_{t-1}$, $a_{t-1}$). 
Specifically, the world model predicts the next state \( (s_t,  \mathcal{G}_t) \) by sampling $N_s$ times from $\gP_{\text{wm}}(s_t, \mathcal{G}_t | s_{t-1}, \mathcal{G}_{t-1}, a_{t-1})$.
Then the discriminator \( \gP_{\text{d}} \) selects the top candidate state.
Based on the selected  \( (s_t,  \mathcal{G}_t) \), the controller determines whether to continue reasoning.
If the goal is achieved, the controller $\gP_{\text{c}}(A|G, s_t,  \mathcal{G}_t)$ generates the final answer \( A \), stores $(G, s_t,  \mathcal{G}_t, A)$ in the completed pool $\mathcal{D}$, and adds the completed count \( b \) by 1.
Otherwise, \( (G, H, s_t,  \mathcal{G}_t) \) will be added to the context pool $\mathcal{C}$ for the next step. 
The process continues until the step limit \( T \) is reached or \( b \) reaches the breadth limit $B$. 
Finally, the top answer \( A^* \) is selected by the discriminator \( \gP_{\text{d}} \) based on the completed states in $\mathcal{D}$.

\subsection{Seeking diversity in action generation}

\begin{figure}[t]
    \centering
    \includegraphics[width=0.75\linewidth]{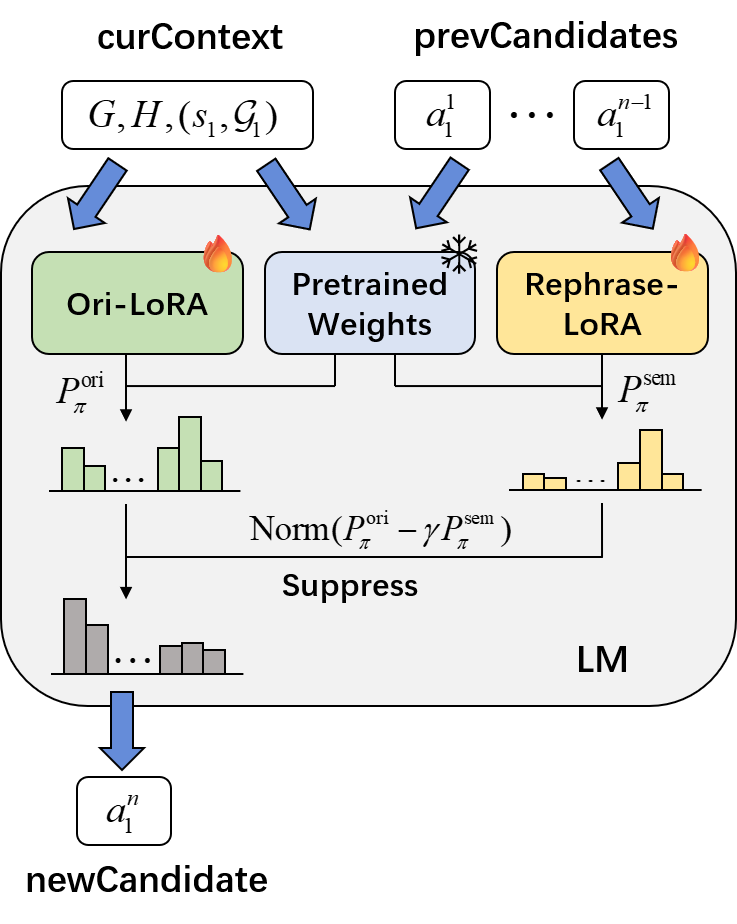}
    \caption{Overview of the Diversity-Based Modeling (DM) method. The original distribution is computed from the current context, and a paraphrasing distribution from previous candidates is subtracted to encourage diversity. We denote each candidate for action 1 as $a_1^j$, where the subscript 1 indicates the action index and the superscript $j \in \{1, \dots, n\}$ denotes the candidate index.}
    \label{fig:DiversityBasedModelling}
\vspace{-10pt}
\end{figure}

We identify two fundamental bottlenecks in planning-based reasoning: \textbf{generation diversity} and \textbf{discrimination accuracy}.
Enhancing diversity is crucial for expanding the solution space, increasing the likelihood of discovering globally optimal solutions. 
To address this, we propose a \textbf{Diversity-based Modelling} (DM) approach (Figure \ref{fig:DiversityBasedModelling}).
The key idea is to encourage the policy to \textbf{generate multiple diverse candidate options for the same step}, ensuring that each new option differs from previous ones, thereby mitigating self-bias and promoting exploration.

Given the current state $(s_t, \mathcal{G}_t)$, we define $\gP^{\text{ori}}_{\pi}\left(a_t | G, H, s_t, \mathcal{G}_t \right)$ as the original distribution learned from positive trajectories during training. 
For $n$-th generation, we aim to introduce diversity by considering an additional distribution $\gP^{\text{sem}}_{\pi}(a^n_t|a_t^{1..n-1})$, which captures steps that are semantically equivalent to those generated previously $a_t^{1..n-1}$. Specifically, the probability of $l$-th token $a^{n}_{t,l}$ in the $n$-th generation $a_t^n$ is
\begin{equation}
\begin{split}
\gP^{\text{sem}}_{\pi}(a^{n}_{t,l}| a_t^{1..n-1}, a^{n}_{t,1..l-1}) = \\
\frac{1}{n-1} \sum_{j=1}^{n-1} 
\gP^{\text{sem}}_{\pi}(a^{n}_{t,l}|a_t^{j}, a^{n}_{t,1..l-1})
\end{split}
\end{equation}
where $a_t^{j}$ denotes the $j$-th generation, and for notational simplicity, we {move the token index to a subscript},
so that $a^{n}_{t,1..l-1}$ denotes the preceding tokens of the $l$-th token $a^{n}_{t,l}$.
The distribution $\gP^{\text{sem}}_{\pi}(a^{\prime}_{l}|a, a^{\prime}_{1..l-1})$ is learned from training data generated by GPT-4o, where each pair $(a, a^{\prime})$ consists of semantically equivalent actions.

The final distribution is given by:
\begin{equation}
\begin{split}
    \gP_{\pi}(a^{n}_{t,l} | G, H, s_t, \mathcal{G}_t, a_t^{1..n-1}, a^{n}_{t,1..l-1}) 
    = \\ \text{Norm}\bigg(\gP^{\text{ori}}_{\pi}(a^{n}_{t,l}|G, H, s_t, \mathcal{G}_t, a^{n}_{t,1..l-1}) \\
    \qquad - \gamma_l \gP^{\text{sem}}_{\pi}(a^{n}_{t,l}| a_t^{1..n-1}, a^{n}_{t,1..l-1})\bigg)
\end{split}
\label{equ:DBM}
\end{equation}
where the decay factor, $\gamma_l = \gamma_0 \cdot \alpha^l$ with $\alpha \leq 1$, is introduced to emphasize diversity in early stages of generation while gradually reducing this effect. This ensures that the deduplication effect is stronger initially to explore different paths but weakens over time to avoid drifting too far from plausible solutions, thereby maintaining accuracy.

The normalization function 
\begin{equation}
    \text{Norm}(\gP) = \frac{\max(\gP, 0)}{\mathbf{1}^\top \max(\gP, 0)}
\end{equation}
is applied to discard negative-valued tokens (that either resemble previous responses or deviate from the intended progression of reasoning) and ensure a diverse and relevant response.
The selection of this function is further discussed in Appendix \ref{appendix:method_detail}.

\subsection{Improving discrimination accuracy in reasoning}
\label{subsec: discimination accuracy}
As highlighted in recent works \citep{huang2023large, chen2024tree}, discrimination accuracy is a critical aspect of planning-based methods.
However, training an effective process reward model (PRM) is challenging, as it reduces each candidate option to a single numerical score, oversimplifying complex decision-making aspects.
Moreover, those raw scores may not be inherently comparable without proper calibration and additional context.

\begin{figure}[!t]
    \centering
    \includegraphics[width=0.9\linewidth]{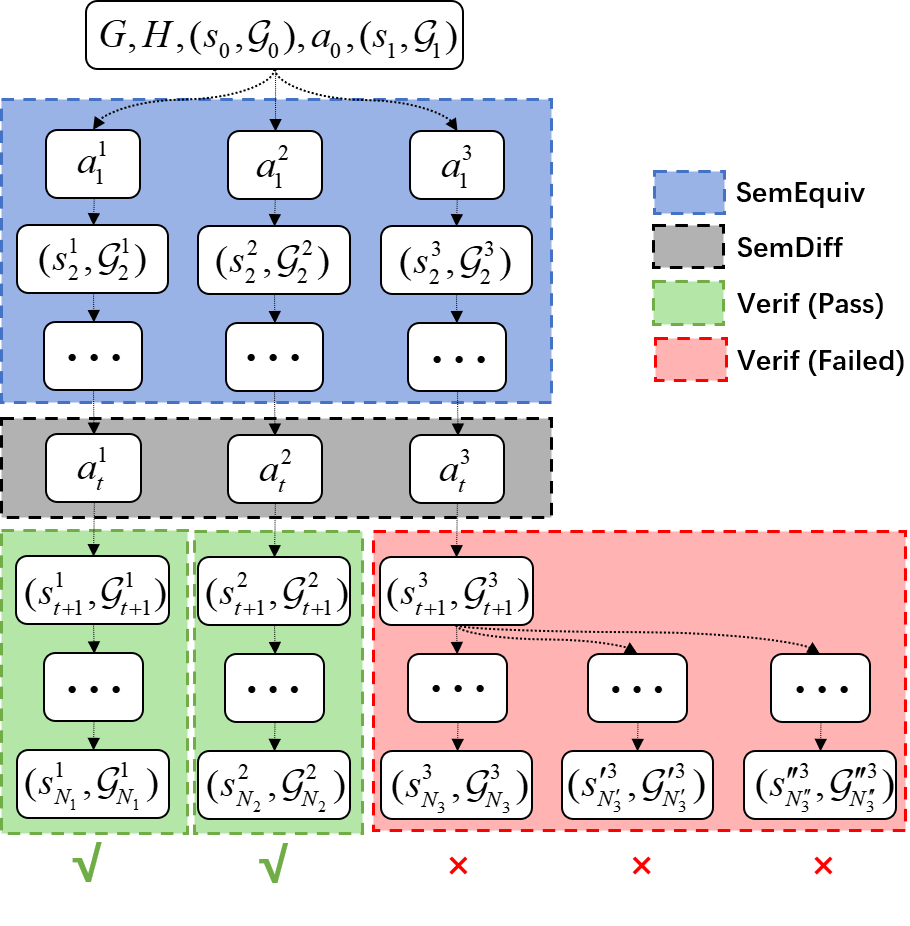}
    \caption{Overview of the automatic annotation process. Beginning from a selected step in the positive trajectory, multiple alternative trajectories are generated. Symbolic verification is first applied to filter out invalid trajectories. Then, the first erroneous steps in negative trajectories are identified through semantic equivalence check and tree-search-based outcome verification.}
    \label{fig:ProcessSupervision}
\vspace{-10pt}
\end{figure}

To address this issue, our discriminator employs \textbf{Contrastive Ranking} (CR) to evaluate multiple candidate options effectively.
By \textbf{directly comparing these options within context}, the discriminator can distinguish nuanced differences, particularly identifying erroneous components, thereby simplifying the evaluation process.
To illustrate the \textbf{automatic annotation process} (Figure \ref{fig:ProcessSupervision}), consider a positive trajectory $\left[(s_0, \mathcal{G}_0), a_0, \cdots, (s_{T}, \mathcal{G}_T)\right]$ that leads to the correct final answer. 
We randomly select an intermediate step $t$, and generate $K$ alternative reasoning trajectories: $\{[a^j_{t}, \cdots, (s^j_{T_j}, \mathcal{G}^j_{T_j})]\}_{j=1}^K$,
where $T_j$ represents the length of the $j$-th trajectory. 
Among these $K$ trajectories, we first filter out invalid trajectories through symbolic verification.
Then the first erroneous steps in negative trajectories are identified by semantically comparing with the corresponding steps in the positive trajectory.
We further perform $N_{\text{veri}}$ completions for {outcome verification}, \ie if none of the completions result in the correct answer, we confirm the identified steps as erroneous.

{
\renewcommand{\arraystretch}{0.95}
\begin{table*}[t]
\centering
\resizebox{0.92\textwidth}{!}{%
\small
\begin{tabular}{l|cc|cc|cc}
\hline
\specialrule{0em}{1pt}{1pt}
& \multicolumn{2}{|c}{Math} & \multicolumn{2}{|c}{Logic} & \multicolumn{2}{|c}{Coding} \\
\cmidrule(lr){2-7}
\multicolumn{1}{l|}{\multirow[c]{-2}{*}{Method}} & \multicolumn{1}{c}{GSM8K} & \multicolumn{1}{c|}{MATH500} & \multicolumn{1}{c}{FOLIO} & \multicolumn{1}{c|}{ReClor} & \multicolumn{1}{c}{HumanEval} & \multicolumn{1}{c}{MBPP} \\
\hline
\specialrule{0em}{1pt}{1pt}
\multicolumn{7}{c}{LLaMA3-8B-Instruct} \\
\hline
\specialrule{0em}{1pt}{1pt}
Zero-shot CoT & 76.9 {\scriptsize{$\pm$ 0.2}} & 27.8 {\scriptsize{$\pm$ 0.1}} & 60.4 {\scriptsize{$\pm$ 0.6}} & 57.8 {\scriptsize{$\pm$ 0.3}} & 60.3 {\scriptsize{$\pm$ 0.3}} & 58.8 {\scriptsize{$\pm$ 0.1}} \\
Few-shot CoT {{(4-shot)}} & 77.6 {\scriptsize{$\pm$ 0.2}} & 25.8 {\scriptsize{$\pm$ 0.3}} & 64.6 {\scriptsize{$\pm$ 0.4}} & 63.9 {\scriptsize{$\pm$ 0.2}} & 59.6 {\scriptsize{$\pm$ 0.1}} & 59.8 {\scriptsize{$\pm$ 0.2}} \\
SFT-CoT & 77.2 {\scriptsize{$\pm$ 0.1}} & 27.0 {\scriptsize{$\pm$ 0.2}} & 66.0 {\scriptsize{$\pm$ 0.4}} & 64.2 {\scriptsize{$\pm$ 0.3}} & 58.6 {\scriptsize{$\pm$ 0.2}} & 60.0 {\scriptsize{$\pm$ 0.2}}\\
SC (CoT@maj32) & 87.3 {\scriptsize{$\pm$ 0.2}} & 30.6 {\scriptsize{$\pm$ 0.3}} & 69.0 {\scriptsize{$\pm$ 0.8}} & 71.1 {\scriptsize{$\pm$ 0.5}} & - & -\\
SC (SFT-CoT@maj32) & 86.4 {\scriptsize{$\pm$ 0.3}} & 28.7 {\scriptsize{$\pm$ 0.3}} & 71.4 {\scriptsize{$\pm$ 0.6}} & 73.2 {\scriptsize{$\pm$ 0.4}} & - & -\\
ToT & 83.0 {\scriptsize{$\pm$ 0.2}} & 25.0 {\scriptsize{$\pm$ 0.3}} & 68.7 {\scriptsize{$\pm$ 0.8}} & 69.0  {\scriptsize{$\pm$ 0.4}} & - & -\\
RAP & 86.2 {\scriptsize{$\pm$ 0.3}} & 26.2 {\scriptsize{$\pm$ 0.3}} & 70.8 {\scriptsize{$\pm$ 0.6}} & 70.4 {\scriptsize{$\pm$ 0.4}} & - & -\\
PRM (PRM800K) & 89.0 {\scriptsize{$\pm$ 0.2}}& 34.6 {\scriptsize{$\pm$ 0.2}}& -& -&- & -\\
PRM (Math-Shepherd) & 90.2 {\scriptsize{$\pm$ 0.1}} & 33.4 {\scriptsize{$\pm$ 0.3}}& -& -&- &- \\
\hline 
\specialrule{0em}{1pt}{1pt}
\methodname{} {(w/o planning)} & 80.1 {\scriptsize{$\pm$ 0.3}} & 32.0 {\scriptsize{$\pm$ 0.2}} & 69.1 {\scriptsize{$\pm$ 0.4}} & 67.4 {\scriptsize{$\pm$ 0.3}}& 62.2 {\scriptsize{$\pm$ 0.2}} & 61.4 {\scriptsize{$\pm$ 0.3}}\\
\methodname{} & \textbf{92.6} {\scriptsize{$\pm$ 0.2}} & \textbf{43.2} {\scriptsize{$\pm$ 0.3}} & \textbf{79.2} {\scriptsize{$\pm$ 0.5}}& \textbf{78.1} {\scriptsize{$\pm$ 0.4}}& \textbf{68.8} {\scriptsize{$\pm$ 0.4}}& \textbf{68.0} {\scriptsize{$\pm$ 0.3}}\\ 
\hline
\specialrule{0em}{1pt}{1pt}
\multicolumn{7}{c}{Mistral-7B-Instruct} \\
\hline
\specialrule{0em}{1pt}{1pt}
Zero-shot CoT & 25.6 {\scriptsize{$\pm$ 0.1}} & 12.8 {\scriptsize{$\pm$ 0.6}} & 49.5 {\scriptsize{$\pm$ 0.6}} & 54.0 {\scriptsize{$\pm$ 0.1}} & 42.3 {\scriptsize{$\pm$ 1.1}} & 38.8 {\scriptsize{$\pm$ 0.4}} \\
Few-shot CoT {{(4-shot)}} & 51.7 {\scriptsize{$\pm$ 0.4}} & 15.0 {\scriptsize{$\pm$ 0.7}} & 57.0 {\scriptsize{$\pm$ 1.1}} & 43.1 {\scriptsize{$\pm$ 0.6}} & 43.6 {\scriptsize{$\pm$ 0.6}} & 44.8 {\scriptsize{$\pm$ 0.6}}\\
SFT-CoT & 52.4 {\scriptsize{$\pm$ 0.2}} & 16.4 {\scriptsize{$\pm$ 0.4}} & 59.0 {\scriptsize{$\pm$ 0.5}} & 56.2 {\scriptsize{$\pm$ 0.2}} & 43.8 {\scriptsize{$\pm$ 0.4}} & 45.8 {\scriptsize{$\pm$ 0.4}}\\
SC (CoT@maj32) & 69.9 {\scriptsize{$\pm$ 0.3}} & 20.6 {\scriptsize{$\pm$ 0.6}} & 61.5 {\scriptsize{$\pm$ 0.7}}& 53.4 {\scriptsize{$\pm$ 0.2}} & -& - \\
SC (SFT-CoT@maj32) & 72.6 {\scriptsize{$\pm$ 0.4}} & 22.3 {\scriptsize{$\pm$ 0.4}} & 64.2 {\scriptsize{$\pm$ 0.6}} & 61.8 {\scriptsize{$\pm$ 0.4}} & - & -\\
ToT & 58.5 {\scriptsize{$\pm$ 0.6}} & 17.3 {\scriptsize{$\pm$ 0.4}} & 58.7 {\scriptsize{$\pm$ 0.6}} & 50.2 {\scriptsize{$\pm$ 0.4}}& - & -\\
RAP & 72.4 {\scriptsize{$\pm$ 0.4}} & 18.8 {\scriptsize{$\pm$ 0.5}} & 62.0 {\scriptsize{$\pm$ 0.4}} & 54.6 {\scriptsize{$\pm$ 0.3}} & - & -\\
PRM (PRM800K) & 74.2 {\scriptsize{$\pm$ 0.3}} & 24.1 {\scriptsize{$\pm$ 0.4}}& -& -& -&- \\
PRM (Math-Shepherd) & 76.0 {\scriptsize{$\pm$ 0.4}}& 22.8 {\scriptsize{$\pm$ 0.6}}& -& -&- &- \\
\hline
\specialrule{0em}{1pt}{1pt}
\methodname{} {({w/o planning})} & 57.0 {\scriptsize{$\pm$ 0.2}} & 19.4 {\scriptsize{$\pm$ 0.4}} & 62.0 {\scriptsize{$\pm$ 0.6}}& 58.2 {\scriptsize{$\pm$ 0.2}} & 45.0 {\scriptsize{$\pm$ 0.3}}& 46.2 {\scriptsize{$\pm$ 0.4}}\\
\methodname{} & \textbf{83.2} {\scriptsize{$\pm$ 0.4}}& \textbf{28.0} {\scriptsize{$\pm$ 0.5}}& \textbf{71.2} {\scriptsize{$\pm$ 0.4}}& \textbf{67.9} {\scriptsize{$\pm$ 0.4}} & \textbf{52.2} {\scriptsize{$\pm$ 0.4}}& \textbf{51.6} {\scriptsize{$\pm$ 0.3}}\\ 
\hline
\end{tabular}
}
\caption{Overall performance comparison across different benchmark datasets for math, logic, and coding tasks. The best performance for each benchmark using the same base model is in bold.}
\label{tab:main-results}
\vspace{-5pt}
\end{table*}
}

Given the process annotations, we define the inputs and outputs of the discriminator while incorporating \textbf{meta knowledge} $\gK_{\text{meta}}$, such as common pitfalls and errors, to improve ranking quality. 
Specifically,
\begin{equation}
\begin{split}
{E}, a_{t}^{\text{best}} \sim \gP_{\text{d}}\Big({E}, a_{t}^{\text{best}} \mid \gK_{\text{meta}}, G, H, s_{t}, \mathcal{G}_t, \\
\{a^{j}_{t}, s^{j}_{t+1}, \mathcal{G}^{j}_{t+1} \}_{j=1}^K \Big)
\end{split}
\end{equation}
\begin{equation}
\resizebox{0.89\hsize}{!}{$
\begin{split}
{E}, s_{t+1}^{\text{best}}, \mathcal{G}_{t+1}^{\text{best}} \sim \gP_{\text{d}} \Big({E}, s_{t+1}^{\text{best}}, \mathcal{G}_{t+1}^{\text{best}} \mid \gK_{\text{meta}}, s_{t}, \mathcal{G}_t, \\
a_{t}, (s_{t+1}^{j}, \mathcal{G}_{t+1}^{j})_{j=1}^K\Big)
\end{split}
$}
\end{equation}
where $E$ denotes an explanation, highlighting differences among the $K$ future states before making a decision.
The superscript 'best' indicates the final selected option, \ie $a_{t}^{\text{best}}$ and $(s_{t+1}^{\text{best}}, \mathcal{G}_{t+1}^{\text{best}})$.
For action ranking, we use the simulated immediate next state $(s_{t+1}, \mathcal{G}_{t+1})$, as $a_t$ only needs to align with the plan $H$.
For plan ranking, we use the simulated terminal state $(s_{T}, \mathcal{G}_{T})$, that is, $\gP_{\text{d}}(E, H^{\text{best}} \mid \gK_{\text{meta}}, G, s_0, \mathcal{G}_0, \{H^{j}, s^{j}_{T_{j}}, \mathcal{G}^{j}_{T_{j}}\}_{j=1}^K)$.
For training data, we select a pair of correct and incorrect options from the positive and negative trajectories of the same question.
We then fine-tune the discriminator using this data, along with meta-knowledge and explanations bootstrapped from GPT-4o.
Further discussions on the methodology and implementation details can be found in Appendix \ref{appendix:method_detail}, \ref{appendix:implementation}.

During {inference}, given a set of candidate options, we apply \textbf{points-based ranking} to select {the top $b$} candidates denoted as \texttt{dis}(\texttt{model}, \texttt{input}, b) (in Algorithm \ref{alg:swap}, \ref{alg:statepredict}). 
To manage computational complexity, we randomly sample comparison groups, ensuring that the total number of comparisons remains \textbf{linear} with respect to the number of candidates.
To further enhance robustness, we randomly reorder group members before evaluation.
Additionally, we perform symbolic verification prior to invoking the discriminator.
Further details on these strategies are provided in Appendix \ref{appendix:method_detail}.

\begin{figure*}[!t]
    \centering
    \includegraphics[width=\linewidth]{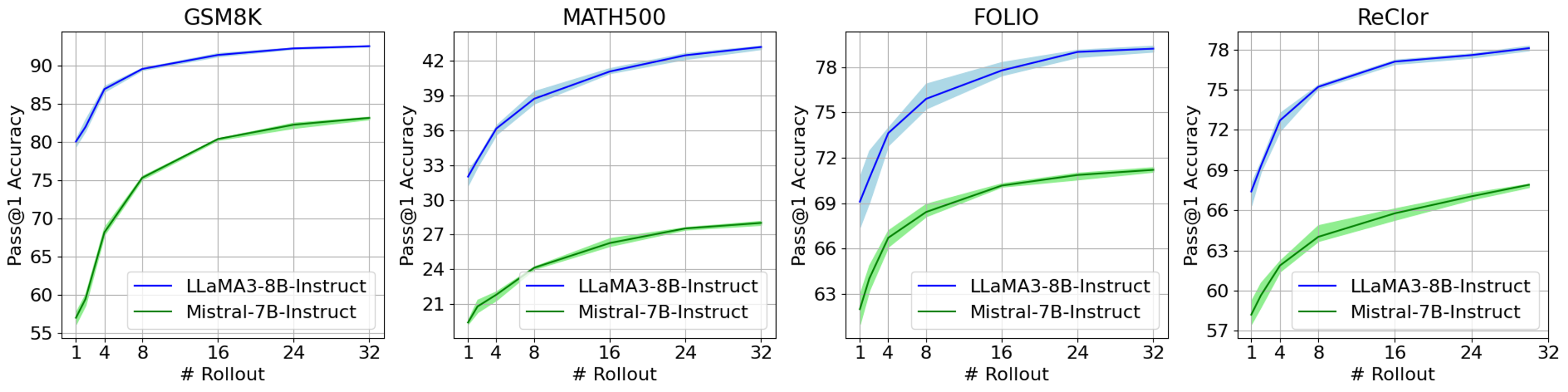}
    \includegraphics[width=\linewidth]{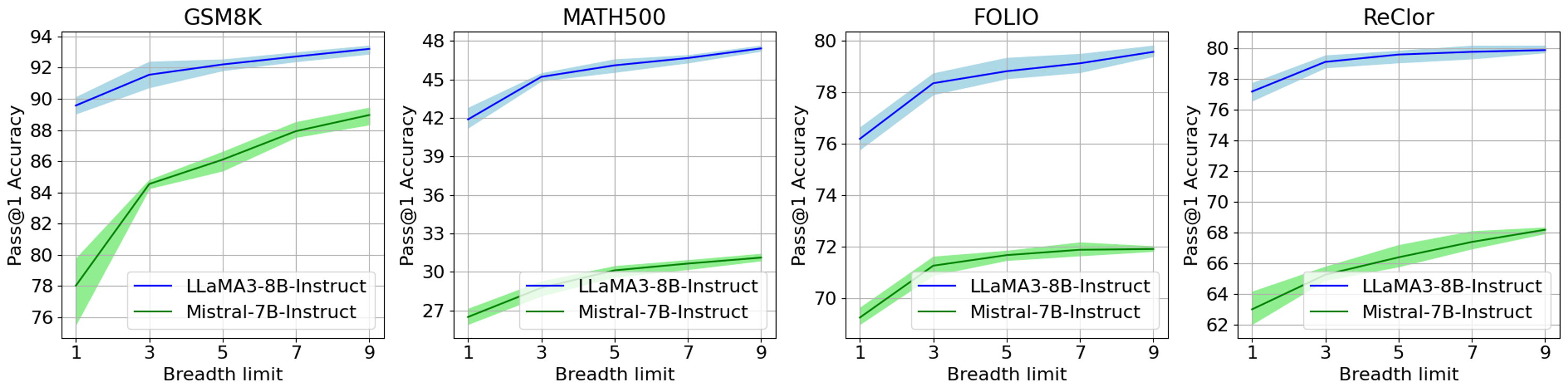}
    \caption{Effect of increasing total rollouts and breadth limit on overall accuracy across benchmarks and base models in \methodname{}. For efficiency, 200 test samples are randomly selected per dataset when breadth is large.}
    \label{fig:EffectofIncreasingTreeWidth}
    \vspace{-5pt}
\end{figure*}

{
\renewcommand{\arraystretch}{1}
\begin{table*}[t]
\centering
\resizebox{0.92\textwidth}{!}{%
\small
\begin{tabular}{l|cc|cc|cc}
\hline 
\specialrule{0em}{1pt}{1pt}
  &\multicolumn{2}{|c}{Math} &\multicolumn{2}{|c}{Logic}&\multicolumn{2}{|c}{Coding} \\
\cmidrule(lr){2-7}
\multicolumn{1}{l|}{\multirow[c]{-2}{*}{Method}}   &\multicolumn{1}{c}{GSM8K} &\multicolumn{1}{c|}{MATH500}&\multicolumn{1}{c}{FOLIO}&\multicolumn{1}{c|}{ReClor} &\multicolumn{1}{c}{HumanEval}&\multicolumn{1}{c}{MBPP}
\\
\hline
\specialrule{0em}{1pt}{1pt}
\methodname{} (Ours) & \textbf{92.6}  & \textbf{43.2}  & \textbf{79.2} & \textbf{78.1} & \textbf{68.8} & \textbf{68.0}  \\
{w/o SFT (ICL only)} & 87.8 & 31.2 & 71.0 & 72.4 & 63.2 & 62.0 \\
{w/o structure info (training)} & 89.0 & 39.9 & 74.6 & 74.0 & 67.0 & 66.0  \\
{w/o structure info (inference)} & 90.2 & 41.6 & 75.8 & 76.2 & 67.4 & 67.0 \\
{w/o state pred refinement} & 88.0 & 38.4 & 76.3 & 74.2 & 67.0 & 66.8\\
{w/o DM} & 87.0 & 36.8 & 74.0 & 75.6 & 65.2 & 64.6 \\
{w/o meta knowledge} & 91.0 & 42.1 & 78.0 & 77.2 & 67.6 & 67.0 \\
{w/o planning} & 80.1  & 32.0  & 69.1  & 67.4 & 62.2  & 61.4 \\
\hline
\end{tabular}
}
\caption{Ablation study results with LLaMA3-8B-Instruct across different benchmarks.}
\label{table-ablation}
\vspace{-10pt}
\end{table*}
}

\vspace{-2pt}
\section{Experiments}
\vspace{-2pt}

\subsection{Experimental setup}
In this section, we demonstrate the versatility and effectiveness of our framework by applying it to a diverse set of reasoning tasks.
Dataset statistics and examples are provided in Appendix \ref{appendix:dataset_overview}.
The key parameter settings in \methodname{} are as follows: 
DM: $\gamma_0 = 0.7$, $\alpha = 0.95$. 
CR: $N_{\text{veri}} = 3$; 
training group size: $K = 2$;
inference group size: less than $3$.
To enhance training effectiveness, we employ advanced training strategies such as curriculum learning and self-improving training for supervised fine-tuning, followed by preference learning for further optimization.
We evaluate \methodname{} against popular reasoning strategies, including: CoT and Self-consistency (SC) \citep{wang2023selfconsistency}, ToT \citep{yao2023tree} and RAP \citep{hao2023reasoning}, Supervised fine-tuning (SFT) on CoTs and verification with PRMs \citep{lightman2023let, wang2023math}.
We conduct experiments with two representative base models: LLaMA3-8B-Instruct \citep{dubey2024llama3herdmodels} and Mistral-7B-Instruct \citep{jiang2023mistral}.
Implementation settings are as follows:
Self-consistency and PRMs use 32 rollouts.
For \methodname{}, ToT, and RAP: 
the generation limit is 3, with a breadth limit of 3.The total number of rollouts is 32. The step limit is 10 for MATH500, and 6 for all other datasets.
Temperature for these methods is set as 0.7. 
More implementation details are provided in Appendix \ref{appendix:datageneration}, \ref{appendix:implementation}.

{
\renewcommand{\arraystretch}{0.95}
\begin{table*}[t]
\centering
\resizebox{0.95\textwidth}{!}{%
\small
\begin{tabular}{l|ccccc}
\hline 
\specialrule{0em}{1pt}{1pt}
 &\multicolumn{1}{c}{Avg. Node} &\multicolumn{1}{c}{Avg. Node}& & &\multicolumn{1}{c}{Final Ans}\\
\multicolumn{1}{l|}{\multirow[c]{-2}{*}{Method}}  &\multicolumn{1}{c}{Content Acc} &\multicolumn{1}{c}{Dependency Acc}& \multicolumn{1}{c}{\multirow[c]{-2}{*}{Avg. Node Acc}} & \multicolumn{1}{c}{\multirow[c]{-2}{*}{Graph Acc}} &\multicolumn{1}{c}{Acc}
\\
\hline
\specialrule{0em}{1pt}{1pt}
Few-shot (4-shot) & 74.6  & 57.2  & 43.8 & 13.0 & 60.0   \\
Few-shot CoT (4-shot) & 88.3 & 80.6 & 77.9 & 47.0 & 68.0  \\
SFT-CoT & 89.2 & 84.4 & 81.8 & 52.0 & 70.0 \\
\methodname{} & \textbf{95.7} & \textbf{91.8} & \textbf{90.0} & \textbf{68.0} & \textbf{81.0}  \\
\hline
\end{tabular}
}
\caption{Graph Construction Evaluation on FOLIO with LLaMA3-8B-Instruct.}
\label{table-graph-eval}
\vspace{-10pt}
\end{table*}
}

\vspace{-2pt}
\subsection{Main results}
\vspace{-2pt}
The overall performance is shown in Table \ref{tab:main-results}, with fine-grained results and example analyses provided in Appendix \ref{appendix: fine_grained_res} and \ref{appendix: examples}, respectively. 
The key findings are summarized as follows:

\noindent \textbf{\methodname{} consistently outperforms all other reasoning methods.} 
Verification approaches, such as self-consistency and PRMs, do not search through intermediate steps during reasoning. 
In contrast, \methodname{} enables the model to engage in structured, conscious planning, similar to human reasoning, which is particularly crucial for avoiding intermediate errors.

\noindent \textbf{Structured representation and an accurate world model further enhance planning effectiveness.} 
Planning-based methods such as ToT and RAP fall short in performance compared to our approach.
They lack the structured understanding and precise state modelling that \methodname{} provides. 
\methodname{} explicitly incorporates structure to capture relationships between key statements, providing priors and enabling symbolic verification. 
Additionally, Diversity-based Modeling (DM) enables the framework to explore a broader solution space, increasing the likelihood of discovering optimal steps.
Meanwhile, Contrastive Ranking (CR) significantly enhances discrimination accuracy by focusing on nuanced differences between candidates.

\vspace{-2pt}
\subsection{Analysis}
\vspace{-2pt}
We examine the impact of total rollouts and breadth limit on overall accuracy (Figure \ref{fig:EffectofIncreasingTreeWidth}), providing insights into optimal parameter selection and demonstrating inference-time scaling.

\noindent \textbf{Multiple rollouts improve accuracy but with diminishing returns.}
Increasing the number of rollouts generally enhances accuracy across all datasets, but the gains taper off as rollouts increase. 
Beyond a certain point, additional rollouts offer minimal improvement, making computational efficiency a key consideration.

\noindent \textbf{Expanding breadth improves accuracy, but with diminishing returns at higher limits.}
Accuracy increases for both models as the breadth limit grows across all datasets, enhancing candidate selection. 
However, the gains are steepest at lower breadth values, with improvements becoming incremental beyond a certain point.

\vspace{-2pt}
\subsection{Graph Construction Evaluation}
Since graph construction is the primary objective of \methodname{}, we conduct an evaluation to further demonstrate its effectiveness. 
However, graph alignment remains a challenge due to node generation and structural variations \cite{dalvi2021explaining}. 
To address this, we manually evaluate different methods on 100 random test samples in FOLIO (Table \ref{table-graph-eval}).
We define our evaluation metrics as follows. 
For each node in a graph, we assess both its content and dependency. 
Node content accuracy (or dependency accuracy) for a single graph is calculated as the proportion of nodes with correct content (or correct dependency). 
We also define node accuracy as the proportion of nodes that are correct in both content and dependency. 
These metrics are averaged across all graphs in the test set. 
A graph is considered fully correct if all its nodes are accurate in both content and dependency.
In addition to our method, we include several baseline methods to highlight the improvements. 
Since our evaluation centers on graph construction, we adapted the output format (the input remains unchanged, but the output includes both the constructed graph and the final answer). 
The graph format adheres to the structure described in \citet{dalvi2021explaining}. 
For CoT-based settings, models are prompted to first produce a chain-of-thought rationale for graph construction, followed by the graph and final answer.

\methodname{} outperforms all baselines across all metrics, demonstrating its effectiveness in generating accurate and structurally correct graphs. 
Notably, \methodname{} achieves significant improvements in both node-level and graph-level accuracy, highlighting the benefits of our structured approach to graph construction. 
Moreover, incorporating CoT significantly improves graph construction, which aligns with our design choice.
We also observed several interesting error patterns in the \methodname{} results. 
For positive trajectories, nodes within shorter contexts (fewer than five premise sentences) are typically entirely correct in both content and dependency. 
However, in longer contexts, occasional errors in dependencies emerge, often involving missing or redundant parent nodes. 
In contrast, for negative trajectories, content errors are more likely to occur in the final conclusion node or in its immediate parent nodes.

\vspace{-2pt}
\subsection{Ablation study}
We analyze the impact of the key components introduced in this paper (Table \ref{table-ablation}). 
The complete framework achieves the highest performance across all tasks, demonstrating that each component positively contributes to overall accuracy. 
Notably, planning has the most significant impact by effectively selecting optimal actions.
Removing supervised fine-tuning leads to notable performance drops, highlighting the value of training on high-quality trajectories. 
Structural information at training and inference contributes to improved learning and generalization. 
State prediction refinement and diversity modeling enhance internal consistency and expand the exploration space, with notable benefits in logic tasks. 
These improvements are consistently observed across different benchmarks, highlighting the robustness of \methodname{}.

\begin{table}[t]
\centering
\small
\renewcommand{\arraystretch}{0.9}
\resizebox{\linewidth}{!}{
\begin{tabular}{l|cc}
\toprule
  & \multicolumn{2}{c}{GSM8K} \\
\cmidrule(lr){2-3}
Method  & Avg. Gen. Tokens & Acc \\
\midrule
\specialrule{0em}{1pt}{1pt}
Zero-shot CoT & 0.17k & 76.9 \\
Few-shot CoT (4-shot) & 0.15k & 77.6 \\
SC (CoT@maj32) & 5.2k & 87.3\\
ToT & 10.8k & 83.0  \\
RAP & 18.6k & 86.2 \\
\methodname{} {({w/o planning})} & 0.31k & 80.1 \\
\methodname{} & 12.4k & 92.6 \\
\bottomrule
\end{tabular}
}
\caption{Efficiency-performance trade-off analysis with LLaMA3-8B-Instruct across different methods.}
\label{tab:effiency_study}
\vspace{-10pt}
\end{table}

\subsection{Efficiency study}
In this section, we analyze the efficiency of different planning methods.
We evaluate the average number of tokens generated by different methods on GSM8K dataset using Llama-3-8B-Instruct.
The results are summarized in Table \ref{tab:effiency_study}. 
We observed that while the theoretical time complexity of \methodname{} is comparable to ToT (BFS with pruning), it generates more tokens in practice due to the incorporation of a world model and the construction of entailment graphs.
On the other hand, \methodname{} is significantly more efficient than RAP, which requires simulations until terminal states to estimate the expected future rewards.
The examples in Appendix~\ref{lst:tot-gsm8k} demonstrate that \methodname{} produces coherent, goal-aligned reasoning with accurate intermediate steps. 
In contrast, ToT often misinterprets key concepts, and RAP yields inconsistent outputs. 
\methodname{} not only improves answer correctness and interpretability but also achieves greater efficiency than RAP by avoiding full trajectory simulations.

\vspace{-2pt}
\section{Related Work}
\vspace{-2pt}
Advanced planning methods for enhancing the multi-step problem-solving capabilities of language models (LMs) can be categorized into three main approaches:
{re-ranking} \citep{wang2023selfconsistency, li-etal-2023-making, lei2024macmutilizingmultiagentcondition}, 
{iterative correction} \citep{madaan2023selfrefine, shinn2023reflexion, yao2022react, chen2024selfdebug} and 
tree search \citep{gu2023don, hao2023reasoning, yao2023tree, zhou2023language}. 
Despite differences in design, these methods fundamentally rely on a discriminator to evaluate planning steps.
Recent studies  \citep{huang2023large, chen2024tree} have demonstrated that the discriminator plays a crucial role in planning-based reasoning.
As a result, using in-context learning to prompt the same LM as both the generator and discriminator is often insufficient for improving model performance on complex reasoning tasks.

To address this limitation, prior research has explored various discriminator (reward model) designs.
There are two primary types of reward models:
Outcome Reward Model (ORM): evaluates the fully generated solution by assigning a single scalar confidence score, trained through outcome supervision by comparing generated answers with the ground truth;
Process Reward Model (PRM): provides stepwise rewards throughout the reasoning process, assigning a scalar confidence score to each intermediate step \citep{lightman2023let, yuan2024advancing, tian2024toward}.
Empirical evidence suggests that process supervision offers greater benefits to multi-step reasoning than outcome supervision, as it ensures the correctness of each intermediate step \citep{lightman2023let}.
However, training a PRM requires process supervision, which is difficult to obtain—manual process annotations are inherently not scalable. 
Although recent research \citep{wang2023math, luo2024improve} has explored automatic process annotation via tree search, training an effective PRM remains challenging.
A key limitation is that PRMs reduce each candidate option to a single numerical score, which oversimplifies complex decision-making.
Moreover, those raw scores may not be inherently comparable without proper calibration and additional context. 

To address these challenges, we introduce Contrastive Ranking (CR), which compares multiple candidate options within prompts rather than evaluating them in isolation.
By directly comparing options, the model can:
distinguish nuanced differences between candidates;
identify erroneous components more effectively;
and simplify the evaluation process, leading to improved discrimination accuracy.
Additionally, we enhance the automatic annotation process by incorporating semantic equivalence checks and symbolic verification.

\vspace{-2pt}
\section{Conclusion}
\vspace{-2pt}

In this paper, we propose  \methodname{}, a novel framework that enhances the reasoning capabilities of LMs through structure-aware planning with an accurate world model. 
Extensive experiments demonstrate that \methodname{} consistently outperforms existing methods, achieving significant improvements across reasoning-intensive benchmarks.
Our approach primarily adopts a re-ranking strategy, which balances computational efficiency and model performance.
For future work, exploring reinforcement learning to enable dynamic interaction between the policy model and the world model, along with tool integration, could further optimize LMs for solving complex, long-horizon real-world problems.
Additionally, teaching models to self-identify and correct mistakes presents a promising direction for enhancing robustness within our framework.

\section*{Limitations}

Planning-based methods, including \methodname{}, enhance deliberative reasoning in language models.
However, generating multiple candidate options and comparing them for each intermediate step can be resource-intensive.
Future work could focus on optimizing the procedure or exploring the trade-offs between accuracy and efficiency.
One promising direction is to develop adaptive strategies that dynamically adjust the number of candidates based on task difficulty or the model’s confidence at each reasoning step.
Additionally, resource-aware algorithms could be explored to explicitly balance computational cost and accuracy.
To further improve efficiency, lightweight model architectures could be leveraged to enhance the discriminator, enabling it to quickly and accurately compare candidates.

\section*{Ethics Statement}

In this paper, we fine-tune language models using publicly available benchmarks intended for research purposes. Additionally, we employ multiple language models to generate text based on these benchmarks. While the authors of these models are committed to addressing ethical considerations, generated content may still contain improper or harmful material. None of such content reflects the opinions of the authors.

\section*{Acknowledgments}
This work is supported in part by DARPA SciFy program, Award No.HR001125C0302, and CISCO Systems, Inc.

\bibliography{custom}

\appendix

\clearpage

\section{Further Methodological Details}
\label{appendix:method_detail}

\noindent \textbf{Normalization function selection.}
The normalization function 
\begin{equation} \notag
    \text{Norm}(\gP) = \frac{\max(\gP, 0)}{\mathbf{1}^\top \max(\gP, 0)}
\end{equation}
is applied to discard negative-valued tokens (that either resemble previous responses or deviate from the intended progression of reasoning) and ensure a diverse and relevant response.

Negative values in the adjusted probability, \ie $\gP^{\text{ori}}_{\pi} - \gamma \gP^{\text{sem}}_{\pi}$, typically arise in two cases:

(1) Repeated tokens: The token’s original probability $\gP^{\text{ori}}_{\pi}$ is high, but its semantic equivalence probability $\gP^{\text{sem}}_{\pi}$ is even higher. 
This suggests that the token closely resembles previous responses, making it less desirable for diversity.

(2) Irrelevant tokens: The token’s original probability $\gP^{\text{ori}}_{\pi}$ is not high, indicating that it likely represents a step deviating from the intended reasoning process.

In both cases, clipping negative values to zero eliminates undesirable tokens while preserving diverse and contextually relevant outputs.
Alternative normalization methods, such as Softmax, redistribute probability mass across all tokens, including those that should ideally have very low or zero probability.
This distorts the original probability distribution, allowing irrelevant tokens to persist and ultimately degrading model performance.

Consider an intermediate step in a math problem with the original probability distribution: 
$\gP^{\text{ori}}_{\pi}$ = \{"Let": 0.6, "Simplify": 0.2, "Consider": 0.2, other tokens: 0\}. 
Given the previous response "Let $x$ be 4.", the semantic equivalence probability distribution becomes:
$\gP^{\text{sem}}_{\pi}$ = \{"Let": 0.5, "Set": 0.2, "Consider": 0.3, other tokens: 0\}. 
Let $\gamma = 1$, then the adjusted probability distribution is:
$\gP^{\text{ori}}_{\pi} - \gamma \gP^{\text{sem}}_{\pi}$ = {"Let": 0.1, "Simplify": 0.2, "Consider": -0.1, "Set": -0.2, other tokens: 0}.
Applying the proposed method yields: 
\{"Let": 0.33, "Simplify": 0.66, every other token: 0\}.
This ensures that only high-relevance tokens are preserved.
However, applying Softmax results in:
\{"Let": 0.00011, "Simplify": 0.00012, "Consider": 0.00009, "Set": 0.00008, other tokens: 0.0001\}.
Here, probability mass is spread thinly across all tokens, diluting the relevance of the original distribution and harming model performance.
Thus, our proposed normalization method effectively filters out irrelevant tokens while preserving diversity and relevance, ensuring a more accurate and controlled reasoning process.

\noindent \textbf{State prediction refinement with diversity.}
Encouraging the world model to generate diverse predictions increases the likelihood of overcoming self-biases and discovering a more accurate future state. 
To ensure robustness, we select the top prediction from the diverse options generated.
To achieve this, we apply a similar strategy to enhance diversity for state prediction, that is,
\begin{equation}
\resizebox{0.89\hsize}{!}{$
\begin{split}
    \gP_{\text{wm}}(s^{n}_{t,l} | s_{t-1}, \mathcal{G}_{t-1}, a_{t-1}, s_{t}^{1..n-1}, s^{n}_{t,1..l-1}) 
    = \\
    \text{Norm}\Big(\gP^{\text{ori}}_{\text{wm}}(s^{n}_{t,l}|s_{t-1}, \mathcal{G}_{t-1}, a_{t-1}, s^{n}_{t,1..l-1})  \\
     - \gamma_l \gP^{\text{sem}}_{\text{wm}}(s^{n}_{t,l}| s_{t}^{1..n-1}, s^{n}_{t,1..l-1})\Big)
\end{split}
$}
\end{equation}
where $s^{j}_{t}$ denotes the $j$-th response, and $s^{n}_{t, 1..l-1}$ is the preceding tokens of the $l$-th token $s^{n}_{t,l}$.
Once the state $s^n_{t}$ is generated, the corresponding graph $\mathcal{G}^n_{t}$ is extracted, ensuring the model maintains a consistent representation of entailment relationships throughout the reasoning process.

\noindent \textbf{Contextual reframing strategy.}
In addition to Diversity-based Modelling (DM), we employ a {contextual reframing strategy} to further enhance generation diversity.
This approach involves randomly reinterpreting the current state to create an alternative context at each step. 
For example, given the original state $(s, \mathcal{G})$, we generate an alternative state $(s^{\prime}, \mathcal{G}^{\prime})$,
where $s^{\prime}$ is sampled from the semantic equivalence distributions \(\gP^{\text{sem}}_{\text{wm}}\left(s^{\prime}| s\right)\). 
The corresponding graph $\mathcal{G}^{\prime}$ is then regenerated from $s^{\prime}$ to maintain consistency in the reasoning process. 
Our experiments demonstrate that this strategy also significantly enhances output diversity, improving the robustness and performance of the model on reasoning tasks.

\noindent \textbf{Comparison with related work.}
Compared to existing approaches \citep{vijayakumar2016diverse, hu2023amortizing}, Diversity-based Modeling (DM) offers several advantages: 
(1) End-to-end learning: DM operates as a post-training strategy for LMs, enabling seamless integration and scalability to large datasets.
(2) Leveraging pre-trained knowledge: It utilizes the extensive world knowledge embedded in pre-trained LMs, enhancing generation diversity without requiring additional domain-specific training.
Diverse beam search \citep{vijayakumar2016diverse} performs beam search in groups using a diversity-augmented objective, but it has several limitations:
(1) Computational complexity: Beam search at the token level becomes computationally intractable for long-form reasoning.
(2) Hyperparameter sensitivity: Identifying optimal strategies and hyperparameters for similarity calculation is time-consuming.
(3) Reliability issues in specialized tasks: In reasoning tasks involving special tokens (e.g., math reasoning or first-order logic reasoning), embedding-based similarity calculations may be unreliable.
GFlowNets fine-tuning \citep{hu2023amortizing} is a diversity-seeking reinforcement learning algorithm based on amortized Bayesian inference.  
Although it demonstrates better performance compared to SFT with limited training data, it is unclear whether it can scale to large-scale datasets and complex reasoning tasks. 
As a reinforcement learning method, GFlowNets fine-tuning can be significantly more challenging and costly to train when dealing with large-scale datasets.

\noindent \textbf{Contrastive ranking for candidate selection.}
During inference, given a set of candidate options, we apply points-based ranking to select the top $b$ candidates denoted as $\texttt{dis}(\texttt{model}, \texttt{input}, b)$.
Specifically, ranking procedure includes:
(1) Comparison group formation:
We consider all possible option combinations for a fixed comparison group size.
To manage computational complexity, we randomly sample comparison groups, ensuring the total number of comparisons remains linear with respect to the number of candidates.
(2) Point assignment:
Within each comparison group, the discriminator selects the best option, which receives 1 point, while the remaining options receive 0 points.
(3) Robustness enhancement:
To reduce positional bias, we randomly reorder group members before evaluation.
(4) Final ranking:
After all comparisons, candidates are ranked based on their total points.
Additionally, before invoking the discriminator, we apply symbolic verification to discard invalid options.
Options that fail symbolic verification are excluded from ranking.

\noindent \textbf{Meta knowledge construction.}
The {meta knowledge} $\gK_{\text{meta}}$, which aids in answer verification and error identification, is derived from training questions.
We employ GPT-4o with in-context learning to extract meta knowledge from training data, focusing on common pitfalls and errors associated with the same question type. 

Formally, meta knowledge is constructed as: 
\begin{equation}
\gK_{\text{meta}} = \text{concat}_{m \in \mathcal{M}}(\gK_m)
\end{equation}
where $\gK_m$ represents stored knowledge from the $m$-th training sample,
\( \mathcal{M} \) represents the top \( M \) training samples, selected based on the cosine similarity between the training query embedding \( \vq_m \) and the test query embedding \( \vq \):  
\begin{equation}
\mathcal{M} = \arg\max_{ |\mathcal{M}| = M} \sum_{m \in \mathcal{M}} \cos(\vq_m, \vq)
\end{equation}
In practice, when $M$ is large, we reduce the context length by employing a LM-based compressor, which condenses the retrieved meta-knowledge into a shorter sequence of embeddings while preserving essential information.

\noindent \textbf{Symbolic verification.}
To further enhance discrimination, we introduce symbolic verification for generated entailment graphs $\mathcal{G}$. 
The verification process consists of the following key steps:
(1) Syntax verification: Ensures that nodes and edges adhere to the correct format.
(2) Node dependency analysis: Examines the dependencies between nodes (given conditions, assumptions, facts, or inferences derived from prior nodes).
(3) Cycle detection: Verifies that the graph remains acyclic, preventing logical contradictions.
(4) Redundancy check: Detects redundant or disconnected nodes.
Each step is implemented using standard graph algorithms, ensuring efficient and reliable verification.

\section{Dataset Overview}
\label{appendix:dataset_overview}
In this section, we provide statistics and examples for all benchmark datasets used in our study. 
We consider GSM8K \citep{cobbe2021gsm8k}, MATH \citep{hendrycksmath2021} for math reasoning, FOLIO \citep{han2022folio}, ReClor \citep{yu2020reclor} for logical reasoning, and HumanEval \citep{chen2021evaluating}, MBPP \citep{austin2021program} for coding.
For GSM8K, there are 7,473 training samples and 1,319 test samples. 
For MATH, there are 7,500 training samples and 5,000 test samples. 
MATH500 is a subset of 500 representative test samples extracted by \citet{lightman2023let}, with the remaining test samples added to the training set.
For FOLIO, the training and validation sets consist of 1,001 and 203 samples, respectively. 
For ReClor, we use 4,638 training samples, 500 validation samples (used as test set, as the original test set answers are not publicly available), and 1,000 test samples. 
HumanEval contains 164 test samples, and since it lacks a training set, we use the entire MBPP dataset (after format transformation) for training. 
MBPP consists of 374 training samples, 90 validation samples, and 500 test samples.
This dataset selection ensures a comprehensive evaluation across diverse reasoning tasks.

\section{Prompts for Data Generation}
\label{appendix:datageneration}

In this section, we present all the prompts used in our data generation process. These prompts include those for plan generation, action generation, state generation, final answer generation, semantic equivalence data generation, semantic equivalence evaluation, meta knowledge generation, and contrastive process supervision for plan, action, and state generation.

\section{Implementation Details}
\label{appendix:implementation}
\noindent \textbf{Training data creation.}
For each dataset, we use multiple  models (GPT-4o \citep{openai2024gpt4technicalreport}, DeepSeek \citep{deepseekai2024deepseekv2strongeconomicalefficient, liu2024deepseek}, LLaMA3 \citep{dubey2024llama3herdmodels}) to generate diverse trajectories for the training and validation sets. 
Before labeling, we filter out trajectories that fail symbolic verification.
We then label trajectories as positive or negative based on their final answers.
To improve model stability, we augment training samples using GPT-4o.
Given the positive and negative trajectories of the same question, we automatically generate process annotations (as discussed in Section \ref{subsec: discimination accuracy}) using DeepSeek.
Additionally, we observed an order bias in contrastive ranking data, where the model tends to prefer the first option when presented with similar choices.
To address this, we apply pre-processing and post-processing techniques:
(1) Pre-processing:
We randomly shuffle the correct option’s position before training.
Only responses that select the correct option are accepted.
(2) Post-processing:
During inference, we randomly reorder options to mitigate bias and improve robustness.
These strategies ensure more stable training and fairer ranking evaluations.

\noindent \textbf{Model training overview.} 
\methodname{} is fine-tuned from a base model using LoRA \citep{hu2021lora}.
The primary objective is to train the LM to understand the format and structure of entailment graphs.
We choose LoRA over full fine-tuning due to the limited availability of training data.
To ensure scalability and generalization, we fine-tune a single generator, which is then repurposed to serve as the policy model, world model, and controller. 
For each dataset, the generator is fine-tuned on all positive trajectories from the training set. 
As illustrated in Figure \ref{fig:DiversityBasedModelling}, the generator contains two LoRAs. 
The original LoRA is fine-tuned on positive trajectories as usual, while the SemEquiv-LoRA is fine-tuned on semantic equivalence data,  which are bootstrapped using GPT-4o, for plan, actions and states.
Specially, the total number of trajectories used for generator fine-tuning: GSM8k (28.3k), MATH500 (49.3k), FOLIO (7.3k), ReClor (14.5k), HumanEval (6.0k), and MBPP (3.4k).
For each positive trajectory, we random sampled some steps and generated two alternatives for each step as semantically equivalent pairs. 
The total number of semantically equivalent pairs obtained:
GSM8k (8.1k), MATH500 (24.2k), FOLIO (3.8k), ReClor (7.1k), HumanEval (2.8k) and MBPP (2.0k).

As a critical component of our framework, the discriminator is fine-tuned from a base model using contrastive ranking data for each dataset.
Specifically, the total number of contrastive ranking pairs used for training:
GSM8k (48.0k), MATH500 (100.2k), FOLIO (14.1k), ReClor (28.7k), HumanEval (10.4k), and MBPP (8.2k).
We explore two approaches for discriminator fine-tuning: full fine-tuning and LoRA.
We found that with sufficient training data, full fine-tuning provides better performance. 
Yet a discriminator fine-tuned with LoRA still yields significant benefits in our framework with a lower computational cost.

\noindent \textbf{Meta knowledge retrieval.} 
We use a DPR model \citep{karpukhin-etal-2020-dense} to generate embeddings for both training questions and the test query.
We then compute cosine similarity to select the top 5 matches during inference.
Once the relevant meta-knowledge is extracted, we explore two approaches:
(1) Using the original text directly.
(2) Compressing the text into a shorter version.
Our experiments show that Approach 1 yields higher accuracy, while Approach 2 offers slightly lower accuracy but faster inference speed.

\noindent \textbf{Future state depth determination.} 
The depth of the future state (mentioned in Section ~\ref{subsec: discimination accuracy}) is determined experimentally.
Our findings indicate that:
For plan ranking, the terminal state is the most effective.
For action ranking, the immediate next state is sufficient, while using deeper future states leads to a decline in accuracy.
After analyzing error cases, we attribute this decline to new errors introduced by subsequent actions, which negatively impact evaluation.

{
\renewcommand{\arraystretch}{0.9}
\begin{table*}[t]
\centering
\small
\begin{tabular}{l|cccccccc}
\hline 
\specialrule{0em}{1pt}{1pt}
  &\multicolumn{8}{|c}{Math500} \\
\cmidrule(lr){2-9}
 &\multicolumn{1}{c}{ALG} &\multicolumn{1}{c}{CP}&\multicolumn{1}{c}{GEO}&\multicolumn{1}{c}{IA} &\multicolumn{1}{c}{NT}&\multicolumn{1}{c}{PRE}&\multicolumn{1}{c}{PALG}&\multicolumn{1}{c}{Total}
\\
\multicolumn{1}{l|}{\multirow[c]{-3}{*}{Method}} &\multicolumn{1}{c}{\scriptsize{(\# 124)}} &\multicolumn{1}{c}{\scriptsize{(\# 38)}}&\multicolumn{1}{c}{\scriptsize{(\# 41)}}&\multicolumn{1}{c}{\scriptsize{(\# 97)}} &\multicolumn{1}{c}{\scriptsize{(\# 62)}}&\multicolumn{1}{c}{\scriptsize{(\# 56)}}&\multicolumn{1}{c}{\scriptsize{(\# 82)}}&\multicolumn{1}{c}{\scriptsize{(\# 500)}}
\\
\hline
\specialrule{0em}{1pt}{1pt}
{Zero-shot CoT} & 46.8 & 21.6&20.0 &7.4 &17.7 & 16.4& 45.1& 27.8\\
{Few-shot CoT (4-shot)} & 41.9&15.3 &22.4 &12.8 &24.5 &11.1 &34.4 &25.8  \\
{SFT-CoT} & 42.9 & 16.7& 21.7 & 12.8 & 24.5 & 12.7 & 38.8 & 27.0 \\
{Self-consistency (@maj32)} & 45.4& 25.7& 25.4& 15.2& 31.4&  9.1& 45.4& 30.6  \\
\methodname{} {(w/o planning)} & 46.8 & 23.0 & 24.6 & 15.9 & 31.8 & 15.8 & 47.8 & 32.0 \\
\methodname{} & \textbf{52.3}& \textbf{49.3}& \textbf{35.6}& \textbf{22.1}& \textbf{43.4}&\textbf{25.9} &\textbf{66.8}&\textbf{43.2}  \\ 
\hline
\end{tabular}
\caption{Fine-grained performance with LLaMA3-8B-Instruct on MATH500 across different subsets: Algebra (ALG), Counting and Probability (CP), Geometry (GEO), Intermediate Algebra (IA), Number Theory (NT), Precalculus (PRE), and Prealgebra (PALG). The number of test questions for each subset is shown in parentheses. Bold values indicate the best performance per subset and overall.}
\label{tab:fine_grained_res_subset}
\end{table*}
}

{
\renewcommand{\arraystretch}{0.9}
\begin{table*}[t]
\centering
\small
\begin{tabular}{l|cccccc}
\hline 
\specialrule{0em}{1pt}{1pt}
  &\multicolumn{6}{|c}{Math500} \\
\cmidrule(lr){2-7}
  &\multicolumn{1}{c}{L1} &\multicolumn{1}{c}{L2} &\multicolumn{1}{c}{L3}&\multicolumn{1}{c}{L4}&\multicolumn{1}{c}{L5}&\multicolumn{1}{c}{Total}
\\
\multicolumn{1}{l|}{\multirow[c]{-3}{*}{Method}}  &\multicolumn{1}{c}{\scriptsize{(\# 43)}} &\multicolumn{1}{c}{\scriptsize{(\# 90)}} &\multicolumn{1}{c}{\scriptsize{(\# 105)}}&\multicolumn{1}{c}{\scriptsize{(\# 128)}}&\multicolumn{1}{c}{\scriptsize{(\# 134)}}&\multicolumn{1}{c}{\scriptsize{(\# 500)}}
\\
\hline
\specialrule{0em}{1pt}{1pt}
{Zero-shot CoT} &74.4 &44.9 &35.4 & 18.0& 4.6&27.8  \\
{Few-shot CoT (4-shot)} & 51.2& 48.9& 29.5&15.3 &9.3 &25.8   \\
{SFT-CoT} & 69.9 & 48.0 & 30.9 & 14.5 & 7.9 & 27.0\\
{Self-consistency (@maj32)} & 67.9& 50.7& 34.5& 24.5&7.7 &  30.6  \\
\methodname{} {(w/o planning)} & 75.6 & 52.4 & 36.3 & 20.1 & 12.4 & 32.0 \\
\methodname{} & \textbf{83.0} & \textbf{68.5} & \textbf{46.6} & \textbf{36.2} & \textbf{17.2} & \textbf{43.2} \\ 
\hline
\end{tabular}
\caption{Fine-grained performance with LLaMA3-8B-Instruct on MATH500 across different difficulty levels (Level 1-5). The number of test questions for each level is shown. Bold values indicate the best performance.}
\label{tab:fine_grained_res_level}
\end{table*}
}

\noindent \textbf{Advanced training strategies.} 
To ensure effective training of the discriminator, we employ specialized strategies, including curriculum learning and self-improving training for supervised fine-tuning. 
We apply curriculum learning primarily to the MATH500 dataset:
Round 1: Train on Level 1 problems.
Round 2: Train on Level 1 and Level 2 problems.
This process continues until all five levels are included.
In each round, we train the model until convergence, using early stopping to prevent overfitting.
To further refine the discriminator’s accuracy, we implement self-improving training:
Run the trained system on the training samples and collect errors made by the generator.
Fine-tune the discriminator using these errors, while keeping the generator fixed. 
Repeat the process until convergence.
We also employ DPO ~\citep{rafailov2024direct} to enhance the discriminator:
Given a prompt, the discriminator generates multiple responses.
We select response pairs that rank different options as the best.
GPT-4o serves as an expert, providing preference labels for these pairs, which are then used as DPO training data.
Note that we keep the generator’s reasoning capability fixed after conventional supervised fine-tuning, as our goal is to highlight the power of planning within our framework.
However, higher overall system performance could be achieved by revisiting and updating the generator as well.

\section{Fine-grained Results}
\label{appendix: fine_grained_res}
To gain a comprehensive understanding of the model's strengths and weaknesses, we provide fine-grained results on MATH500 (Table \ref{tab:fine_grained_res_subset} and \ref{tab:fine_grained_res_level}). 
We choose MATH500 for this analysis since it categorizes the test set by both problem types and difficulty levels, facilitating a more detailed evaluation of model performance across different dimensions. 
From Table \ref{tab:fine_grained_res_subset}, we observe that \methodname{} consistently outperforms other methods across all subsets and overall. 
This demonstrates that \methodname{} significantly enhances the overall mathematical reasoning capability compared to the baselines. 
The inclusion of the planning mechanism enables more accurate reasoning and selection, improving performance across different subsets.
Similarly, in Table \ref{tab:fine_grained_res_level}, \methodname{} achieved the best performance across all difficulty levels, particularly excelling in the most challenging Level 5. 
The planning mechanism contributes to improved accuracy on high-difficulty problems, demonstrating its effectiveness in enhancing reasoning capabilities. 
As difficulty increases, all methods show a significant decline in performance, particularly at Levels 4 and 5, indicating the increased complexity of reasoning required for these problems. 
Overall, \methodname{} consistently outperforms the baseline, especially on higher-difficulty problems, highlighting its advantage in handling complex reasoning tasks.

\section{Output Examples of \methodname{}}
\label{appendix: examples}

In this section, we provide example outputs generated using LLaMA3-8B-Instruct with \methodname{} for all benchmarks used in our paper, including GSM8K, MATH500, FOLIO, ReClor, HumanEval, and MBPP.

\onecolumn
\begin{lstlisting}[title=Example - GSM8K]
$\textbf{Problem:}$ Weng earns $\$$12 an hour for babysitting. Yesterday, she just did 50 minutes of babysitting. How much did she earn?

$\textbf{Solution:}$ 
Weng earns 12/60 = $\$$<<12/60=0.2>>0.2 per minute.
Working 50 minutes, she earned 0.2 x 50 = $\$$<<0.2*50=10>>10.
#### 10

$\textbf{Problem:}$ Janet hires six employees. Four of them are warehouse workers who make $\$$15/hour, and the other two are managers who make $\$$20/hour. Janet has to pay 10% of her workers' salaries in FICA taxes. If everyone works 25 days a month and 8 hours a day, how much does Janet owe total for their wages and taxes for one month?

$\textbf{Solution:}$ 
First figure out how many hours each worker works per month by multiplying the number of days they work by the number of hours a day they work: 25 days * 8 hours/day = <<25*8=200>>200 hours
Then calculate how much one warehouse worker makes per month by multiplying their hourly rate by the number of hours they work: 200 hours * $\$$15/hour = $\$$<<200*15=3000>>3000
Then multiply that number by 4 to find out how much all the warehouse workers make: $\$$3000/worker * 4 workers = $\$$<<3000*4=12000>>12,000
Now multiply the hours each manager works (also 200) by their hourly wage to find out how much one manager makes per month: 200 hours * $\$$20/hour = $\$$<<200*20=4000>>4,000
Now multiply one manager's wages by the number of managers (2) to find their total wage amount: $\$$4,000/manager * 2 managers = $\$$<<4000*2=8000>>8,000
Now add the wages for the managers and the workers to find the total cost of the wages: $\$$8,000 + $\$$12,000 = $\$$<<8000+12000=20000>>20,000
Now multiply the total wage bill by 10% to find how much the FICA taxes are: $\$$20,000 * .1 = $\$$<<20000*.1=2000>>2,000
Now add the total wage bill to the total tax amount to find the grand total: $\$$2,000 + $\$$20,000 = $\$$<<2000+20000=22000>>22,000
#### 22000
\end{lstlisting}

\begin{lstlisting}[title=Example - MATH500]
$\textbf{Problem:}$ Let $f(x) = \left\{
\begin{array}{cl} 
ax+3, &\text{ if } x > 2, \\
x-5 &\text{ if } -2 \le x \le 2, \\
2x - b &\text{ if } x < -2.
\end{array}
\right.
$
Find $a+b$ if the piecewise function is continuous (which means that its graph can be drawn without lifting your pencil from the paper).

$\textbf{Solution:}$
For the piecewise function to be continuous, the cases must "meet" at $2$ and $-2$. For example, $ax+3$ and $x-5$ must be equal when $x=2$. This implies $a(2)+3=2-5$, which we solve to get $2a=-6 \Rightarrow a=-3$. Similarly, $x-5$ and $2x-b$ must be equal when $x=-2$. Substituting, we get $-2-5=2(-2)-b$, which implies $b=3$. So $a+b=-3+3=\boxed{0}$.

$\textbf{Problem:}$ Square ABCD has its center at $(8,-8)$ and has an area of 4 square units. The top side of the square is horizontal. The square is then dilated with the dilation center at (0,0) and a scale factor of 2. What are the coordinates of the vertex of the image of square ABCD that is farthest from the origin? Give your answer as an ordered pair.

$\textbf{Solution:}$
With the center of dilation at the origin and a scale factor of 2, all the coordinates of square $ABCD$ are twice the coordinates of its preimage. The preimage has an area of 4 square units, so its side length is 2 units. Since the center of the preimage is at $(8, -8)$, the four vertices of the preimage are at $(7, -9), (7, -7), (9, -7)$ and $(9, -9)$. The point $(9, -9)$ is the farthest from the origin on the preimage, so the point farthest from the origin on the image of square $ABCD$ is $\boxed{(18, -18)}.$
\end{lstlisting}

\begin{lstlisting}[title=Example - FOLIO]
$\textbf{Problem:}$
Premises: 
All customers in James' family who subscribe to AMC A-List are eligible to watch three movies every week without any additional fees. 
Some of the customers in James' family go to the cinema every week. 
Customers in James' family subscribe to AMC A-List or HBO service. 
Customers in James' family who prefer TV series will not watch TV series in cinemas. 
All customers in James' family who subscribe to HBO services prefer TV series to movies. 
Lily is in James' family; she watches TV series in cinemas.

Conclusion:
Lily goes to cinemas every week or watches 3 movies every week without any additional fees.

$\textbf{Solution:}$ True

$\textbf{Problem:}$
Premises:
If a legislator is found guilty of stealing government funds, they will be suspended from office. 
Tiffany T. Alston was a legislator in Maryland's House of Delegates from 2011 to 2013. 
Tiffany T. Alston was found guilty of stealing government funds in 2012.

Conclusion:
Tiffany T. Alston went to prison for stealing government funds.

$\textbf{Solution:}$ Uncertain
\end{lstlisting}

\begin{lstlisting}[title=Example - ReClor]
$\textbf{Problem:}$
Paula will visit the dentist tomorrow morning only if Bill goes golfing in the morning. Bill will not go golfing unless Damien agrees to go golfing too. However, Damien has decided not to go golfing. Ttherefore, Paula will not be visiting the dentist tomorrow morning.

The pattern of reasoning displayed above most closely parallels which of the following?

0. If Marge goes to the bank today, Lauren will not cash her check tomorrow. Marge will not wash her car unless it is sunny. However, it is sunny, so Marge will wash her car and go shopping with Lauren.
1. Kevin will wash his car tomorrow only if Brittany has to go visit her grandmother. Unless Aunt Susan has to run errands, Brittany will not have to go visit her grandmother. Since Aunt Susan does not have to run errands, Kevin will not wash his car tomorrow.
2. Renee will do her homework tonight if there is nothing good on television and if her neighbors do not have a party. Alth- ough, there is something good on television; her neighbors are also having a party. Ttherefore, Renee will attend the party.
3. Maddie will plan a picnic only if one of her friends, Lisa or Kenny, will come. Kenny will not come to the picnic, but Lisa will. Ttherefore, Maddie will plan a picnic.

$\textbf{Solution:}$ 1

$\textbf{Problem:}$
Environmentalist: An increased number of oil spills and the consequent damage to the environment indicate the need for stricter safety standards for the oil industry. Since the industry refuses to take action, it is the national government that must regulate industry safety standards. In particular, the government has to at least require oil companies to put double hulls on their tankers and to assume financial responsibility for accidents. Industry representative: The industry alone should be responsible for devising safety standards because of its expertise in handling oil and its understanding of the cost entailed. Implementing the double-hull proposal is not currently feasible because it creates new safety issues. Furthermore, the cost would be burdensome to the industry and consumers.

Which one of the following is an assumption on which the argument of the environmentalist depends?

0. Government safety regulations are developed in a process of negotiation with industry leaders and independent experts.
1. Environmental concerns outweigh all financial considerations when developing safety standards.
2. The requirement of two hulls on oil tankers, although initially costly, will save money over time by reducing cleanup costs.
3. The only effective sources of increased stringency in safety standards for oil tankers are action by the industry itself or national government regulation.

$\textbf{Solution:}$ 3
\end{lstlisting}

\begin{lstlisting}[title=Example - HumanEval]
$\textbf{Problem:}$
from typing import List


def separate_paren_groups(paren_string: str) -> List[str]:
    ''' Input to this function is a string containing multiple groups of nested parentheses. Your goal is to
    separate those group into separate strings and return the list of those.
    Separate groups are balanced (each open brace is properly closed) and not nested within each other
    Ignore any spaces in the input string.
    >>> separate_paren_groups('( ) (( )) (( )( ))')
    ['()', '(())', '(()())']
    '''

$\textbf{Solution:}$
    result = []
    current_string = []
    current_depth = 0
    
    for c in paren_string:
        if c == '(':
            current_depth += 1
            current_string.append(c)
        elif c == ')':
            current_depth -= 1
            current_string.append(c)

            if current_depth == 0:
                result.append(''.join(current_string))
                current_string.clear()
    
    return result


$\textbf{Problem:}$
def count_up_to(n):
    ''' Implement a function that takes an non-negative integer and returns an array of the first n
    integers that are prime numbers and less than n.
    for example:
    count_up_to(5) => [2,3]
    count_up_to(11) => [2,3,5,7]
    count_up_to(0) => []
    count_up_to(20) => [2,3,5,7,11,13,17,19]
    count_up_to(1) => []
    count_up_to(18) => [2,3,5,7,11,13,17]
    '''

$\textbf{Solution:}$
    primes = []
    for i in range(2, n):
        is_prime = True
        for j in range(2, i):
            if i % j == 0:
                is_prime = False
                break
        if is_prime:
            primes.append(i)
    return primes
\end{lstlisting}

\begin{lstlisting}[title=Example - MBPP]
$\textbf{Problem:}$
Write a function to find the minimum cost path to reach (m, n) from (0, 0) for the given cost matrix cost[][] and a position (m, n) in cost[][].

assert min_cost([[1, 2, 3], [4, 8, 2], [1, 5, 3]], 2, 2) == 8
assert min_cost([[2, 3, 4], [5, 9, 3], [2, 6, 4]], 2, 2) == 12
assert min_cost([[3, 4, 5], [6, 10, 4], [3, 7, 5]], 2, 2) == 16

$\textbf{Solution:}$
R = 3
C = 3
def min_cost(cost, m, n): 
    tc = [[0 for x in range(C)] for x in range(R)] 
    tc[0][0] = cost[0][0] 
    for i in range(1, m+1): 
        tc[i][0] = tc[i-1][0] + cost[i][0] 
    for j in range(1, n+1): 
        tc[0][j] = tc[0][j-1] + cost[0][j] 
    for i in range(1, m+1): 
        for j in range(1, n+1): 
            tc[i][j] = min(tc[i-1][j-1], tc[i-1][j], tc[i][j-1]) + cost[i][j] 
    return tc[m][n]


$\textbf{Problem:}$
Write a function to count the longest repeating subsequences such that the two subsequences don't have same string characters at same positions.

assert find_longest_repeating_subseq(''AABEBCDD'') == 3
assert find_longest_repeating_subseq(''aabb'') == 2
assert find_longest_repeating_subseq(''aab'') == 1

$\textbf{Solution:}$
def find_longest_repeating_subseq(str): 
    n = len(str) 
    dp = [[0 for k in range(n+1)] for l in range(n+1)] 
    for i in range(1, n+1): 
    for j in range(1, n+1): 
        if (str[i-1] == str[j-1] and i != j): 
            dp[i][j] = 1 + dp[i-1][j-1] 
        else: 
            dp[i][j] = max(dp[i][j-1], dp[i-1][j]) 
    return dp[n][n]
\end{lstlisting}

\begin{lstlisting}[title=Prompt - Plan Generation]
Based on the goal, and the initial state (including the graph), propose a plan. Do not solve the problem; just outline the steps for proceeding. 

Example:
### Input:
''Problem'': ''Solve for $a$: $\sqrt{4+\sqrt{16+16a}} + \sqrt{1+\sqrt{1+a}} = 6$.''
''Goal'': ''Solve $a$.''
''Initial state'': ''We know that $ \sqrt{4+\sqrt{16+16a}} + \sqrt{1+\sqrt{1+a}} = 6 $.''
''Initial graph'': {''Statement'': {''s1'': ''$ \sqrt{4+\sqrt{16+16a}} + \sqrt{1+\sqrt{1+a}} = 6 $''}, ''Entailment'': {''s1'': ''Given condition''}}

### Output:
''Plan'': ''To solve $a$, we begin by simplifying $\sqrt{4+\sqrt{16+16a}}$. This simplification may also help us simplify the left side of the equation further.''
\end{lstlisting}

\begin{lstlisting}[title=Prompt - Action Generation]
Based on the goal, the plan, and the history of actions and states (including graphs), propose the next action. Only specify the action itself; do not provide the outcome.

Example 1:
### Input:
''Problem'': ''Solve for $a$: $\sqrt{4+\sqrt{16+16a}} + \sqrt{1+\sqrt{1+a}} = 6$.''
''Goal'': ''Solve $a$.''
''Initial state'': ''We know that $\sqrt{4+\sqrt{16+16a}} + \sqrt{1+\sqrt{1+a}} = 6$.''
''Initial graph'': {''Statement'': {''s1'': ''$ \sqrt{4+\sqrt{16+16a}} + \sqrt{1+\sqrt{1+a}} = 6 $''}, ''Entailment'': {''s1'': ''Given condition''}}
''Plan'': ''To solve $a$, we begin by simplifying $\sqrt{4+\sqrt{16+16a}}$. This simplification may also help us simplify the left side of the equation further.''

### Output:
''Action 1'': ''Simplify $ \sqrt{4+\sqrt{16+16a}} $.''


Example 2:
### Input:
''Problem'': ''Solve for $a$: $\sqrt{4+\sqrt{16+16a}} + \sqrt{1+\sqrt{1+a}} = 6$.''
''Goal'': ''Solve $a$.''
''Initial state'': ''We know that $ \sqrt{4+\sqrt{16+16a}} + \sqrt{1+\sqrt{1+a}} = 6 $.''
''Initial graph'': {''Statement'': {''s1'': ''$ \sqrt{4+\sqrt{16+16a}} + \sqrt{1+\sqrt{1+a}} = 6 $''}, ''Entailment'': {''s1'': ''Given condition''}}
''Plan'': ''To solve $a$, we begin by simplifying $\sqrt{4+\sqrt{16+16a}}$. This simplification may also help us simplify the left side of the equation further.''
''Action 1'': ''Simplify $ \sqrt{4+\sqrt{16+16a}} $.''
''State 1'': ''We start by simplifying the expression under the square root.\n\nFirst, observe that:\n$16 + 16a = 16(1 + a)$\nSubstitute this back:\n$\sqrt{4 + \sqrt{16(1 + a)}}$\n\nNow, simplify $\sqrt{16(1 + a)}$:\n$\sqrt{16(1 + a)} = 4\sqrt{1 + a}$\n\nSo the expression becomes:\n$\sqrt{4 + 4\sqrt{1 + a}}$\n\nNext, factor out the 4 from inside the square root:\n$\sqrt{4(1 + \sqrt{1 + a})} $\n\nTaking the square root of 4, we get 2:\n$2\sqrt{1 + \sqrt{1 + a}}$\nSo, the simplified form of the expression is:\n$2\sqrt{1 + \sqrt{1 + a}}$.''
''Graph 1'': {''Statement'': {''s2'': ''$ \sqrt{4+\sqrt{16+16a}} = 2\sqrt{1+\sqrt{1+a}} $''}, ''Entailment'': {''s2'': ''Fact''}}

### Output:
''Action 2'': ''Combine like terms $ \sqrt{1 + \sqrt{1 + a}} $.''
\end{lstlisting}

\begin{lstlisting}[title=Prompt - State Generation]
Based on the goal, the plan, and the history of actions and states (including graphs), predict the next state. Additionally, extract the graph that illustrates key statements and their entailment relations. 

Example 1:
### Input:
''Problem'': ''Solve for $a$: $\sqrt{4+\sqrt{16+16a}} + \sqrt{1+\sqrt{1+a}} = 6$.''
''Goal'': ''Solve $a$.''
''Initial state'': ''We know that $ \sqrt{4+\sqrt{16+16a}} + \sqrt{1+\sqrt{1+a}} = 6 $.''
''Initial graph'': {''Statement'': {''s1'': ''$ \sqrt{4+\sqrt{16+16a}} + \sqrt{1+\sqrt{1+a}} = 6 $''}, ''Entailment'': {''s1'': ''Given condition''}}
''Plan'': ''To solve $a$, we begin by simplifying $\sqrt{4+\sqrt{16+16a}}$. This simplification may also help us simplify the left side of the equation further.''
''Action 1'': ''Simplify $ \sqrt{4+\sqrt{16+16a}} $.''

### Output:
''State 1'': ''We start by simplifying the expression under the square root.\n\nFirst, observe that:\n$16 + 16a = 16(1 + a)$\nSubstitute this back:\n$\sqrt{4 + \sqrt{16(1 + a)}}$\n\nNow, simplify $\sqrt{16(1 + a)}$:\n$\sqrt{16(1 + a)} = 4\sqrt{1 + a}$\n\nSo the expression becomes:\n$\sqrt{4 + 4\sqrt{1 + a}}$\n\nNext, factor out the 4 from inside the square root:\n$\sqrt{4(1 + \sqrt{1 + a})} $\n\nTaking the square root of 4, we get 2:\n$2\sqrt{1 + \sqrt{1 + a}}$\nSo, the simplified form of the expression is:\n$2\sqrt{1 + \sqrt{1 + a}}$.''
''Graph 1'': {''Statement'': {''s2'': ''$ \sqrt{4+\sqrt{16+16a}} = 2\sqrt{1+\sqrt{1+a}} $''}, ''Entailment'': {''s2'': ''Fact''}}


Example 2:
### Input:
''Problem'': ''Solve for $a$: $\sqrt{4+\sqrt{16+16a}} + \sqrt{1+\sqrt{1+a}} = 6$.''
''Goal'': ''Solve $a$.''
''Initial state'': ''We know that $ \sqrt{4+\sqrt{16+16a}} + \sqrt{1+\sqrt{1+a}} = 6 $.''
''Initial graph'': {''Statement'': {''s1'': ''$ \sqrt{4+\sqrt{16+16a}} + \sqrt{1+\sqrt{1+a}} = 6 $''}, ''Entailment'': {''s1'': ''Given condition''}}
''Plan'': ''To solve $a$, we begin by simplifying $\sqrt{4+\sqrt{16+16a}}$. This simplification may also help us simplify the left side of the equation further.''
''Action 1'': ''Simplify $ \sqrt{4+\sqrt{16+16a}} $.''
''State 1'': ''We start by simplifying the expression under the square root.\n\nFirst, observe that:\n$16 + 16a = 16(1 + a)$\nSubstitute this back:\n$\sqrt{4 + \sqrt{16(1 + a)}}$\n\nNow, simplify $\sqrt{16(1 + a)}$:\n$\sqrt{16(1 + a)} = 4\sqrt{1 + a}$\n\nSo the expression becomes:\n$\sqrt{4 + 4\sqrt{1 + a}}$\n\nNext, factor out the 4 from inside the square root:\n$\sqrt{4(1 + \sqrt{1 + a})} $\n\nTaking the square root of 4, we get 2:\n$2\sqrt{1 + \sqrt{1 + a}}$\nSo, the simplified form of the expression is:\n$2\sqrt{1 + \sqrt{1 + a}}$.''
''Graph 1'': {''Statement'': {''s2'': ''$ \sqrt{4+\sqrt{16+16a}} = 2\sqrt{1+\sqrt{1+a}} $''}, ''Entailment'': {''s2'': ''Fact''}}
''Action 2'': ''Combine like terms $ \sqrt{1 + \sqrt{1 + a}} $.''

### Output:
''State 2'': ''We can obtain $\sqrt{4+\sqrt{16+16a}} + \sqrt{1+\sqrt{1+a}} = 2\sqrt{1 + \sqrt{1 + a}} + \sqrt{1+\sqrt{1+a}} $ $ = 3\sqrt{1+\sqrt{1+a}}$.''
''Graph 2'': {''Statement'': {''s3'': ''$ \sqrt{4+\sqrt{16+16a}} + \sqrt{1+\sqrt{1+a}} = 3\sqrt{1+\sqrt{1+a}} $''}, ''Entailment'': {''s3'': [''s1'', ''s2'']}}
\end{lstlisting}

\begin{lstlisting}[title=Prompt - Final Answer Generation]
Based on the goal and the current state (including the graph), determine if the goal has been achieved. If it has, generate the final answer; otherwise, return ''Not yet''.

Example 1:
### Input:
''Problem'': ''Solve for $a$: $\sqrt{4+\sqrt{16+16a}} + \sqrt{1+\sqrt{1+a}} = 6$.''
''Goal'': ''Solve $a$.''
''Initial state'': ''We know that $ \sqrt{4+\sqrt{16+16a}} + \sqrt{1+\sqrt{1+a}} = 6 $.''
''Initial graph'': {''Statement'': {''s1'': ''$ \sqrt{4+\sqrt{16+16a}} + \sqrt{1+\sqrt{1+a}} = 6 $''}, ''Entailment'': {''s1'': ''Given condition''}}
''Plan'': ''To solve $a$, we begin by simplifying $\sqrt{4+\sqrt{16+16a}}$. This simplification may also help us simplify the left side of the equation further.''
''Action 1'': ''Simplify $ \sqrt{4+\sqrt{16+16a}} $.''
''State 1'': ''We start by simplifying the expression under the square root.\n\nFirst, observe that:\n$16 + 16a = 16(1 + a)$\nSubstitute this back:\n$\sqrt{4 + \sqrt{16(1 + a)}}$\n\nNow, simplify $\sqrt{16(1 + a)}$:\n$\sqrt{16(1 + a)} = 4\sqrt{1 + a}$\n\nSo the expression becomes:\n$\sqrt{4 + 4\sqrt{1 + a}}$\n\nNext, factor out the 4 from inside the square root:\n$\sqrt{4(1 + \sqrt{1 + a})} $\n\nTaking the square root of 4, we get 2:\n$2\sqrt{1 + \sqrt{1 + a}}$\nSo, the simplified form of the expression is:\n$2\sqrt{1 + \sqrt{1 + a}}$.''
''Graph 1'': {''Statement'': {''s2'': ''$ \sqrt{4+\sqrt{16+16a}} = 2\sqrt{1+\sqrt{1+a}} $''}, ''Entailment'': {''s2'': ''Fact''}}
''Action 2'': ''Combine like terms $ \sqrt{1 + \sqrt{1 + a}} $.''
''State 2'': ''We can obtain $\sqrt{4+\sqrt{16+16a}} + \sqrt{1+\sqrt{1+a}} = 2\sqrt{1 + \sqrt{1 + a}} + \sqrt{1+\sqrt{1+a}} $ $ = 3\sqrt{1+\sqrt{1+a}}$.''
''Graph 2'': {''Statement'': {''s3'': ''$ \sqrt{4+\sqrt{16+16a}} + \sqrt{1+\sqrt{1+a}} = 3\sqrt{1+\sqrt{1+a}} $''}, ''Entailment'': {''s3'': [''s1'', ''s2'']}}

### Output:
''Not yet''


Example 2:
### Input:
''Problem'': ''Solve for $a$: $\sqrt{4+\sqrt{16+16a}} + \sqrt{1+\sqrt{1+a}} = 6$.''
''Goal'': ''Solve $a$.''
''Initial state'': ''We know that $ \sqrt{4+\sqrt{16+16a}} + \sqrt{1+\sqrt{1+a}} = 6 $.''
''Initial graph'': {''Statement'': {''s1'': ''$ \sqrt{4+\sqrt{16+16a}} + \sqrt{1+\sqrt{1+a}} = 6 $''}, ''Entailment'': {''s1'': ''Given condition''}}
''Plan'': ''To solve $a$, we begin by simplifying $\sqrt{4+\sqrt{16+16a}}$. This simplification may also help us simplify the left side of the equation further.''
''Action 1'': ''Simplify $ \sqrt{4+\sqrt{16+16a}} $.''
''State 1'': ''We start by simplifying the expression under the square root.\n\nFirst, observe that:\n$16 + 16a = 16(1 + a)$\nSubstitute this back:\n$\sqrt{4 + \sqrt{16(1 + a)}}$\n\nNow, simplify $\sqrt{16(1 + a)}$:\n$\sqrt{16(1 + a)} = 4\sqrt{1 + a}$\n\nSo the expression becomes:\n$\sqrt{4 + 4\sqrt{1 + a}}$\n\nNext, factor out the 4 from inside the square root:\n$\sqrt{4(1 + \sqrt{1 + a})} $\n\nTaking the square root of 4, we get 2:\n$2\sqrt{1 + \sqrt{1 + a}}$\nSo, the simplified form of the expression is:\n$2\sqrt{1 + \sqrt{1 + a}}$.''
''Graph 1'': {''Statement'': {''s2'': ''$ \sqrt{4+\sqrt{16+16a}} = 2\sqrt{1+\sqrt{1+a}} $''}, ''Entailment'': {''s2'': ''Fact''}}
''Action 2'': ''Combine like terms $ \sqrt{1 + \sqrt{1 + a}} $.''
''State 2'': ''We can obtain $\sqrt{4+\sqrt{16+16a}} + \sqrt{1+\sqrt{1+a}} = 2\sqrt{1 + \sqrt{1 + a}} + \sqrt{1+\sqrt{1+a}} $ $ = 3\sqrt{1+\sqrt{1+a}}$.''
''Graph 2'': {''Statement'': {''s3'': ''$ \sqrt{4+\sqrt{16+16a}} + \sqrt{1+\sqrt{1+a}} = 3\sqrt{1+\sqrt{1+a}} $''}, ''Entailment'': {''s3'': [''s1'', ''s2'']}}
''Action 3'': ''Solve $a$.''
''State 3'': ''Isolate the square root term by dividing both sides by 3:\n$\sqrt{1+\sqrt{1+a}} = 2$\n\nSquare both sides: $\left( \sqrt{1+\sqrt{1+a}} \right)^2 =$ $2^2$\n$1 + \sqrt{1+a} = 4$\n\nIsolate the inner square root:\n$\sqrt{1+a} = 4 - 1$\n$\sqrt{1+a} = 3$\n\nSquare both sides again:\n$\left( \sqrt{1+a} \right)^2 = 3^2$\n$1 + a = 9$\n\nSolve for $a$:\n$a = 9 - 1$\n$a = 8$\n\nThe solution is $a = 8$.''
''Graph 3'': {''Statement'': {''s4'': ''$ \sqrt{1+\sqrt{1+a}} = 2 $'', ''s5'': ''$ a = 8 $''}, ''Entailment'': {''s4'': [''s1'', ''s3''], ''s5'': [''s4'']}}

### Output:
''Final answer'': ''8''
\end{lstlisting}

\begin{lstlisting}[title=Prompt - Semantic Equivalence Data Generation]
Rewrite the given sentence into two or three different versions. Ensure that each version is distinct in wording and structure. Provide both a thought process and a final answer. In the thought process, include as many details as possible, ensuring that no steps are omitted.

Example 1:
### Input:
''Sentence'': ''$\frac{XZ}{XY} = \frac{ZY}{XY} = \frac{1}{2}.$''

### Output:
''Thought'': ''There are various ways to rewrite the given sentence: 1. $\frac{XZ}{XY} = \frac{1}{2}$ and $\frac{ZY}{XY} = \frac{1}{2}$. 2. $XZ = ZY$ $= \frac{1}{2}XY$. These paraphrased versions capture the same relationship as the original equation.'',
''Answer'': [''$\frac{XZ}{XY} = \frac{1}{2}$ and $\frac{ZY}{XY} = \frac{1}{2}$.'', ''$XZ = ZY = \frac{1}{2}XY$.'']

Example 2:
### Input:
''Sentence'': ''$125a$ is a three-digit number.''

### Output:
''Thought'': ''To rewrite the sentence '$125a$ is a three-digit number', we need to express it in different forms that convey the same meaning. The given sentence implies that $125a$ is within the range of three-digit numbers, i.e., between 100 and 999. Let's break down and reframe the statement as follows: 1. $100 \leq 125a \leq 999$. 2. 'The value of $125a$ lies within the interval [100, 999]'. 3. '$125a$ is an integer that satisfies the condition $100 \leq 125a \leq 999$'. Each of these reframed versions captures the same relationship as the original sentence.''
''Answer'': [''$100 \leq 125a \leq 999$.'', ''The value of $125a$ lies within the interval [100, 999]'', ''$125a$ is an integer that satisfies the condition $100 \leq 125a \leq 999$.'']
\end{lstlisting}

\begin{lstlisting}[title=Prompt - Semantic Equivalence Evaluation]
Compare the provided candidate options, considering both their current attributes and potential future outcomes (if applicable). Determine whether they are semantically equivalent, and respond with either ''same'' or ''different''.

Example:
### Input:
''Problem'': ''Given the polynomials $p(x) = 2x + 3$ and $q(x) = x^2 - x + 4$, find the polynomial resulting from multiplying $p(x)$ and $q(x)$ and express it in standard polynomial form. What is the coefficient of $x^2$ in the resulting polynomial?''
''Goal'': ''Find the coefficient of $x^2$ in the resulting polynomial from multiplying $p(x)$ and $q(x)$.'',
''Initial state'': ''We have the polynomials $p(x) = 2x + 3$ and $q(x) = x^2 - x + 4$.'',
''Initial graph'': {''Statement'': {''s1'': ''$p(x) = 2x + 3$'', ''s2'': ''$q(x) = x^2 - x + 4$''}, ''Entailment'': {''s1'': ''Given condition'', ''s2'': ''Given condition''}},
''Plan'': ''First, multiply the polynomials $p(x)$ and $q(x)$. Then, identify the coefficient of $x^2$ in the resulting polynomial.'',
''Action 1'': ''Multiply the polynomials $p(x)$ and $q(x)$.'',
''State 1'': ''To multiply $p(x) = 2x + 3$ and $q(x) = x^2 - x + 4$, distribute each term of $p(x)$ to each term of $q(x)$: $(2x + 3)(x^2 - x + 4) = 2x(x^2) + 2x(-x) + 2x(4) + 3(x^2) + 3(-x) + 3(4) = 2x^3 - 2x^2 + 8x + $ $3x^2 - 3x + 12$.'',
''Graph 1'': {''Statement'': {''s3'': ''$p(x)q(x) = 2x^3 - 2x^2 + 8x + 3x^2 - 3x + 12$''}, ''Entailment'': {''s3'': [''s1'', ''s2'']}},
''Action 2'': ''Combine like terms in the resulting polynomial.'',
''State 2'': {
    ''Option 1'': ''$2x^3 + (2x^2 - 2x^2) + 8x + 3x^2 + (-3x) + 12 = 2x^3 + 3x^2 + 5x + $ $12$'',
    ''Option 2'': ''$2x^3 + (-2x^2 + 3x^2) + (8x - 3x) + 12 = 2x^3 + x^2 + 5x +$ $12$'',
}
''Future'': {
    ''Future 1'': {''Graph 2'': {''Statement'': {''s4'': ''$p(x)q(x) = 2x^3 + 3x^2 + 5x + 12$''}, ''Entailment'': {''s4'': [''s3'']}}},
    ''Future 2'': {''Graph 2'': {''Statement'': {''s4'': ''$p(x)q(x) = 2x^3 + x^2 + 5x + 12$''}, ''Entailment'': {''s4'': [''s3'']}}}
}

### Output:
''Different''
\end{lstlisting}

\begin{lstlisting}[title=Prompt - Meta Knowledge Generation]
Provide the general knowledge needed to solve the problem. Do not provide a specific solution plan. Instead, offer knowledge that can help verify answers and identify errors.

Example 1:
### Input:
''Problem'': ''Consider the quadratic equation $ax^2 + bx + c = 0$ where $a$, $b$, and $c$ are integers. For this equation to have integer solutions, the discriminant $b^2 - 4ac$ must be a perfect square. Suppose the quadratic equation $2x^2 + kx + 3 = 0$ has integer solutions. What is the value of $k$?''

### Output:
''General knowledge'': ''To avoid errors when determining the value of variables in algebraic equations, follow these general guidelines: 1. Understand conditions for solutions: Know the conditions required for specific solution types and how they relate to equation parameters.  2. Discriminant analysis: Use the discriminant to determine solution types and ensure it meets conditions for the desired type.  3. Factorization techniques: Use factorization to simplify equations and find relationships between parameters.  4. Systematic approach to finding solutions: Solve related equations systematically and consider all potential factor pairs.  5. Verification: Verify solutions by substituting back to confirm correctness and catch errors.''

Example 2:
### Input:
''Problem'': ''Given the polynomials $p(x) = 2x + 3$ and $q(x) = x^2 - x + 4$, find the polynomial resulting from multiplying $p(x)$ and $q(x)$ and express it in standard polynomial form. What is the coefficient of $x^2$ in the resulting polynomial?''

### Output:
''General knowledge'': ''To avoid mistakes when solving polynomial multiplication problems, follow these general guidelines: 1. Understand Polynomial Terms and Their Degrees:  Recognize the individual terms of a polynomial and their respective degrees. 2. Distributive Property in Polynomial Multiplication: Apply the distributive property correctly by multiplying each term of the first polynomial by each term of the second polynomial. 3. Combine Like Terms:  After distributing, combine the like terms, which are terms with the same degree. Be systematic in organizing terms to ensure all like terms are combined correctly. 4. Pay Attention to Signs:  Be careful with positive and negative signs during multiplication and when combining like terms. Ensure that the signs of the terms are handled correctly during the distribution process.''
\end{lstlisting}

\begin{lstlisting}[title=Prompt - Contrastive Process Supervision for Plan Generation]
Compare the provided candidate options, considering both their current attributes and potential future outcomes (if applicable). Pay attention to the meta knowledge. First, present a detailed comparison, showing every step without skipping any. Then, provide a conclusion, selecting only one answer.

Example:
### Input:
''Meta knowledge'': ''To avoid errors when determining the value of variables in algebraic equations, follow these general guidelines: 1. Understand conditions for solutions: Know the conditions required for specific solution types and how they relate to equation parameters.  2. Discriminant analysis: Use the discriminant to determine solution types and ensure it meets conditions for the desired type.  3. Factorization techniques: Use factorization to simplify equations and find relationships between parameters.  4. Systematic approach to finding solutions: Solve related equations systematically and consider all potential factor pairs.  5. Verification: Verify solutions by substituting back to confirm correctness and catch errors.'',
''Problem'': ''Consider the quadratic equation $ax^2 + bx + c = 0$ where $a$, $b$, and $c$ are integers. For this equation to have integer solutions, the discriminant $b^2 - 4ac$ must be a perfect square. Suppose the quadratic equation $2x^2 + kx + 3 = 0$ has integer solutions. What is the value of $k$?'',
''Goal'': ''Find the value of $k$.'',
''Initial state'': ''We know that $2x^2 + kx + 3 = 0$ has integer solutions.'',
''Initial graph'': {''Statement'': {''s1'': ''$2x^2 + kx + 3 = 0$ has integer solutions.''}, ''Entailment'': {''s1'': ''Given condition''}},
''Plan'': {
    ''Option 1'': ''We can use the fact that the discriminant of a quadratic equation must be a perfect square to write an equation in terms of $k$ and then solve for $k$.'',
    ''Option 2'': ''The discriminant $b^2 - 4ac$ must be a perfect square. We will set up the discriminant and solve for $k$ under the condition.''
},
''Future'': {
    ''Future 1'': {''Action 1'': ''Find the discriminant.'', ''State 1'': ''The discriminant is $b^2 - 4ac$, so in this case, it is $k^2 - 24$.'', ''Graph 1'': {''Statement'': {''s2'': ''The discriminant is $k^2 - 24$.''}, ''Entailment'': {''s2'': [''s1'']}}, ''Action 2'': ''Since the discriminant must be a perfect square, we can set $k^2 - 24 = x^2$ for some integer $x$.'', ''State 2'': ''Now, we need to find $k$ by solving the equation $k^2 - 24 = x^2$. We can do this by completing the square: $k^2 - 24 = x^2$. Add $24$ to both sides: $k^2 = x^2 + 24$. Take the square root of both sides: $k = \pm\sqrt{x^2 + 24}$.'', ''Graph 2'': {''Statement'': {''s3'': ''$k = \pm\sqrt{x^2 + 24}$''}, ''Entailment'': {''s3'': [''s1'', ''s2'']}}, ''Action 3'': ''Since $k$ is an integer, $x^2 + 24$ must also be a perfect square. Let $x^2 + 24 = y^2$ for some integer $y$. Then, $k = \pm y$.'', ''State 3'': ''The only values of $y$ that work are $y = \pm4$, so $k = \pm4$. The final answer is $k = 4$ or $k = -4$.'', ''Graph 3'': {''Statement'': {''s4'': ''$k = \pm4$''}, ''Entailment'': {''s4'': [''s3'']}}},
    ''Future 2'': {''Action 1'': ''Calculate the discriminant $b^2 - 4ac$ for the quadratic equation.'', ''State 1'': ''The discriminant $\Delta$ for the equation is given by: $\Delta = k^2 - 4 \cdot 2 \cdot 3 = k^2 - 24$.'', ''Graph 1'': {''Statement'': {''s2'': ''$\Delta = k^2 - 24$.'', ''Entailment'': {''s2'': [''s1'']}}, ''Action 2'': ''Set the discriminant $k^2 - 24$ to be a perfect square.'', ''State 2'': ''For $k^2 - 24$ to be a perfect square, let $k^2 - 24 = n^2$ for some integer $n$. This gives us the equation: $k^2 - n^2 = 24$ $(k - n)(k + n) = 24$.'', ''Graph 2'': {''Statement'': {''s3'': ''$(k - n)(k + n) = 24$''}, ''Entailment'': {''s3'': [''s1'', ''s2'']}}, ''Action 3'': ''Solve for $k$ by finding pairs $(k - n)$ and $(k + n)$ that multiply to 24.'', ''State 3'': ''We need to find integer pairs $(k - n)$ and $(k + n)$ such that their product is 24. Possible pairs are: $(1, 24), (2, 12), (3, 8), (4, 6), (-1, -24), (-2, -12), (-3, -8), (-4, -6)$. Solving for $k$ in each case, we find: $k - n = 2, k + n = 12$, $2k = 14$, $k = 7$.'', ''Graph 3'': {''Statement'': {''s4'': ''$k = 7$''}, ''Entailment'': {''s4'': [''s3'']}}}
}

### Output:
''Comparison'': ''Both options mention that the discriminant must be a perfect square. Based on this observation, they will solve for $k$. Both candidates are valid. However, the futures are different. In Future 1, $k^2 = x^2 + 24$ leads to $k = 4$ or $k = -4$, which is incorrect. Future 2 is correct.'',
''Conclusion'': ''Option 2''
\end{lstlisting} 

\begin{lstlisting}[title=Prompt - Contrastive Process Supervision for Action Generation]
Compare the provided candidate options, considering both their current attributes and potential future outcomes (if applicable). Pay attention to the meta knowledge. First, present a detailed comparison, showing every step without skipping any. Then, provide a conclusion, selecting only one answer.

Example:
### Input:
''Meta knowledge'': ''To avoid errors when determining the value of variables in algebraic equations, follow these general guidelines: 1. Understand conditions for solutions: Know the conditions required for specific solution types and how they relate to equation parameters.  2. Discriminant analysis: Use the discriminant to determine solution types and ensure it meets conditions for the desired type.  3. Factorization techniques: Use factorization to simplify equations and find relationships between parameters.  4. Systematic approach to finding solutions: Solve related equations systematically and consider all potential factor pairs.  5. Verification: Verify solutions by substituting back to confirm correctness and catch errors.'',
''Problem'': ''Consider the quadratic equation $ax^2 + bx + c = 0$ where $a$, $b$, and $c$ are integers. For this equation to have integer solutions, the discriminant $b^2 - 4ac$ must be a perfect square. Suppose the quadratic equation $2x^2 + kx + 3 = 0$ has integer solutions. What is the value of $k$?'',
''Goal'': ''Find the value of $k$.'',
''Initial state'': ''We know that $2x^2 + kx + 3 = 0$ has integer solutions.'',
''Initial graph'': {''Statement'': {''s1'': ''$2x^2 + kx + 3 = 0$ has integer solutions.''}, ''Entailment'': {''s1'': ''Given condition''}},
''Plan'': ''The discriminant $b^2 - 4ac$ must be a perfect square. We will set up the discriminant and solve for $k$ under the condition.'',
''Action 1'': ''Calculate the discriminant $b^2 - 4ac$ for the quadratic equation.'', ''State 1'': ''The discriminant $\Delta$ for the equation is given by: $\Delta = k^2 - 4 \cdot 2 \cdot 3 = k^2 - 24$.'', 
''Graph 1'': {''Statement'': {''s2'': ''$\Delta = k^2 - 24$.'', ''Entailment'': {''s2'': [''s1'']}}, 
''Action 2'': ''Set the discriminant $k^2 - 24$ to be a perfect square.'', 
''State 2'': ''For $k^2 - 24$ to be a perfect square, let $k^2 - 24 = n^2$ for some integer $n$. This gives us the equation: $k^2 - n^2 = 24$ $(k - n)(k + n) = 24$.'',
''Graph 2'': {''Statement'': {''s3'': ''$(k - n)(k + n) = 24$''}, ''Entailment'': {''s3'': [''s1'', ''s2'']}},
''Action 3'': {
  ''Option 1'': ''Solve for $k$ by finding pairs $(k - n)$ and $(k + n)$ that multiply to 24.'',
  ''Option 2'': ''Factor the right side of the equation: $(k - n)(k + n) = 24$.''
},
''Future'': {
    ''Future 1'': {''State 3'': ''We need to find integer pairs $(k - n)$ and $(k + n)$ such that their product is 24. Possible pairs are: $(1, 24), (2, 12), (3, 8), (4, 6), (-1, -24), (-2, -12), (-3, -8), (-4, -6)$. Solving for $k$ in each case, we find: $k - n = 2$, $k + n = 12$, $2k = 14$, $k = 7$.'', ''Graph 3'': {''Statement'': {''s4'': ''$k = 7$''}, ''Entailment'': {''s4'': [''s3'']}}},
    ''Future 2'': {''State 3'': ''The factors of 24 are 1, 2, 3, 4, 6, 8, 12, and 24. We can check each of these to see if they can be expressed as the product of two integers with a sum of $k$ and a difference of $n$. $1 \cdot 24 = 24$, $2 \cdot 12 = 24$, and $3 \cdot 8 = 24$ are the only possibilities. Now, let's check these: $k - n = 1$, $k + n = 24$, $k = 12, n = 23$, $k - n = 2$, $k + n = 12$, $k = 7, n = 5$, and $k - n = 3$, $k + n = 8$, $k = 5, n = 5$. So, the possible values of $k$ are 5, 7, and 12. However, we need to check if $k$ is an integer.'', ''Graph 3'': {''Statement'': {''s4'': ''$k = 5,7,12$''}, ''Entailment'': {''s4'': [''s3'']}}}
}

### Output:
''Comparison'': ''Both options involve solving for $k$ based on the equation. Option 1 mentions finding pairs. Option 2 mentions factorization. Both candidates are valid. Given the actions, the future states are different. Future 1 shows $k = 7$, which is correct. Future 2 shows $k = 5, 7, 12$. In Future 2, by solving $k - n = 1$, $k + n = 24$, we obtain $k = 12.5, n = 11.5$ rather than $k = 12, n = 23$. Thus, Future 2 is incorrect.'',
''Conclusion'': ''Option 1''
\end{lstlisting}

\begin{lstlisting}[title=Prompt - Contrastive Process Supervision for State Generation]
Compare the provided candidate options, considering both their current attributes and potential future outcomes (if applicable). Pay attention to the meta knowledge. First, present a detailed comparison, showing every step without skipping any. Then, provide a conclusion, selecting only one answer.

Example:
### Input:
''Meta knowledge'': ''To avoid mistakes when solving polynomial multiplication problems, follow these general guidelines: 1. Understand Polynomial Terms and Their Degrees:  Recognize the individual terms of a polynomial and their respective degrees. 2. Distributive Property in Polynomial Multiplication: Apply the distributive property correctly by multiplying each term of the first polynomial by each term of the second polynomial. 3. Combine Like Terms:  After distributing, combine the like terms, which are terms with the same degree. Be systematic in organizing terms to ensure all like terms are combined correctly. 4. Pay Attention to Signs:  Be careful with positive and negative signs during multiplication and when combining like terms. Ensure that the signs of the terms are handled correctly during the distribution process.'',
''Problem'': ''Given the polynomials $p(x) = 2x + 3$ and $q(x) = x^2 - x + 4$, find the polynomial resulting from multiplying $p(x)$ and $q(x)$ and express it in standard polynomial form. What is the coefficient of $x^2$ in the resulting polynomial?'',
''Goal'': ''Find the coefficient of $x^2$ in the resulting polynomial from multiplying $p(x)$ and $q(x)$.'',
''Initial state'': ''We have the polynomials $p(x) = 2x + 3$ and $q(x) = x^2 - x + 4$.'',
''Initial graph'': {''Statement'': {''s1'': ''$p(x) = 2x + 3$'', ''s2'': ''$q(x) = x^2 - x + 4$''}, ''Entailment'': {''s1'': ''Given condition'', ''s2'': ''Given condition''}},
''Plan'': ''First, multiply the polynomials $p(x)$ and $q(x)$. Then, identify the coefficient of $x^2$ in the resulting polynomial.'',
''Action 1'': ''Multiply the polynomials $p(x)$ and $q(x)$.'',
''State 1'': ''To multiply $p(x) = 2x + 3$ and $q(x) = x^2 - x + 4$, distribute each term of $p(x)$ to each term of $q(x)$: $(2x + 3)(x^2 - x + 4) = 2x(x^2) + 2x(-x) + 2x(4) + 3(x^2) + 3(-x) + 3(4) = 2x^3 - 2x^2 + 8x + $ $3x^2 - 3x + 12$.'',
''Graph 1'': {''Statement'': {''s3'': ''$p(x)q(x) = 2x^3 - 2x^2 + 8x + 3x^2 - 3x + 12$''}, ''Entailment'': {''s3'': [''s1'', ''s2'']}},
''Action 2'': ''Combine like terms in the resulting polynomial.'',
''State 2'': {
    ''Option 1'': ''$2x^3 + (2x^2 - 2x^2) + 8x + 3x^2 + (-3x) + 12 = 2x^3 + 3x^2 + 5x + $ $12$'',
    ''Option 2'': ''$2x^3 + (-2x^2 + 3x^2) + (8x - 3x) + 12 = 2x^3 + x^2 + 5x +$ $12$''
},
''Future'': {
    ''Future 1'': {''Graph 2'': {''Statement'': {''s4'': ''$p(x)q(x) = 2x^3 + 3x^2 + 5x + 12$''}, ''Entailment'': {''s4'': [''s3'']}}},
    ''Future 2'': {''Graph 2'': {''Statement'': {''s4'': ''$p(x)q(x) = 2x^3 + x^2 + 5x + 12$''}, ''Entailment'': {''s4'': [''s3'']}}}
}

### Output:
''Comparison'': ''Both options involve combining like terms to find the coefficient of $x^2$. Option 1 combines terms resulting in $2x^3 + 3x^2 + 5x + 12$, and Option 2 simplifies to $2x^3 + x^2 + 5x + 12$. Thus, they yield different coefficients for $x^2$ and $x$. Option 1 mentions $2x^3 + (2x^2 - 2x^2) + 8x + $ $3x^2 + (-3x) + 12$. However, State 1 gives $2x^3 - 2x^2 + 8x + 3x^2 - 3x + 12$. Option 1 mistakes $-2x^2$ as $2x^2 - 2x^2$. On the other hand, Option 2 uses the correct equation $2x^3 + (-2x^2 + 3x^2) + (8x - 3x) + 12$, which leads to the correct results.'',
''Conclusion'': ''Option 2''
\end{lstlisting}

\begin{lstlisting}[title=SWAP Output - GSM8K]
$\textbf{Problem:}$ Josh decides to try flipping a house. He buys a house for $\$$80,000 and then puts in $\$$50,000 in repairs. This increased the value of the house by 150%. How much profit did he make?

$\textbf{Solution:}$ 
''Goal'': ''Determine how much profit Josh made after flipping the house.'' 
''Initial state'': ''We know that Josh bought a house for $\$$80,000 and put in $\$$50,000 in repairs, increasing the value by 150%.''
''Initial graph'': {''Statement'': {''s1'': ''Josh bought a house for $\$$80,000.'', ''s2'': ''Josh put in $\$$50,000 in repairs.'', ''s3'': ''Josh increased the value by 150%.''}, ''Entailment'': {''s1'': ''Given condition'', ''s2'': ''Given condition'', ''s3'': ''Given condition''}} 
''Plan'': ''First, calculate the total investment made by adding the purchase price and repair costs. Then calculate the increased value of the house by multiplying the original value by 1.5. Finally, subtract the total investment from the new value to determine the profit.'' 
''Action 1'': ''Calculate the total investment by adding $\$$80,000 (purchase price) and $\$$50,000 (repair costs).'' 
''State 1'': ''Total investment is $\$$80,000 + $\$$50,000 = $\$$130,000.'' 
''Graph 1'': {''Statement'': {''s4'': ''Total investment is $\$$130,000.''}, ''Entailment'': {''s4'': [''s1'', ''s2'']}}
''Action 2'': ''Calculate the increased value of the house by multiplying $\$$80,000 by 1.5.'' 
''State 2'': ''The increase in value is $\$$80,000 * 1.5 = $\$$120,000.'' 
''Graph 2'': {''Statement'': {''s5'': ''The increase in value is $\$$120,000.''}, ''Entailment'': {''s5'': [''s1'', ''s3'']}}
''Action 3'': ''Calculate the new value of the house by adding $\$$80,000 and $\$$120,000.'' 
''State 3'': ''The new value of the house is $\$$80,000 + $\$$120,000 = $\$$200,000.'' 
''Graph 3'': {''Statement'': {''s6'': ''The new value of the house is $\$$200,000.''}, ''Entailment'': {''s6'': [''s1'', ''s5'']}}
''Action 4'': ''Subtract the total investment ($\$$130,000) from the new value of the house ($\$$200,000) to determine the profit.'' 
''State 4'': ''The profit is $\$$200,000 - $\$$130,000 = $\$$70,000.'' 
''Graph 4'': {''Statement'': {''s7'': ''The profit is $\$$70,000.''}, ''Entailment'': {''s7'': [''s4'', ''s6'']}}
''Final answer'': ''70000''
\end{lstlisting}

\begin{lstlisting}[title=SWAP Output - MATH500]
$\textbf{Problem:}$ Find the distance between the vertex of the graph of the equation $f(x) = x^2 - 8x + 15$ and the point $(0, 2)$.

$\textbf{Solution:}$
''Goal'': ''Find the distance between the vertex of the parabola and the point $(0, 2)$.''
''Initial state'': ''We need to find the vertex of the parabola defined by the function $f(x) = x^2 - 8x + 15$.''
''Initial graph'': {''Statement'': {''s1'': ''The function is $f(x) = x^2 - 8x + 15$.'', ''s2'': ''The point is $(0, 2)$.''}, ''Entailment'': {''s1'': ''Given condition'', ''s2'': ''Given condition''}}
''Plan'': ''To find the vertex of the parabola, we can use the formula for the vertex of a quadratic function in the form $f(x) = ax^2 + bx + c$. Then, we calculate the distance between this vertex and the point $(0, 2)$.''
''Action 1'': ''Find the vertex of the quadratic function $f(x) = x^2 - 8x + 15$.''
''State 1'': ''The vertex of a quadratic function $f(x) = ax^2 + bx + c$ is given by the formula $x = -\frac{b}{2a}$. For $f(x) = x^2 - 8x + 15$:\n$a = 1$, $b = -8$, and $c = 15$.\nSubstituting these values:\n$x = -\frac{-8}{2 \cdot 1} = \frac{8}{2} = 4$\n\nTo find the $y$-coordinate of the vertex, substitute $x = 4$ back into $f(x)$:\n$f(4) = 4^2 - 8 \times 4 + 15 = 16 - 32 + 15 = -1$\nThus, the vertex of the parabola is at $(4, -1)$.''
''Graph 1'': {''Statement'': {''s3'': ''The vertex of $f(x) = ax^2 + bx + c$ is given by $x = -\frac{b}{2a}$.'', ''s4'': ''The vertex of the function is $(4, -1)$.''}, ''Entailment'': {''s3'': ''Fact'', ''s4'': [''s1'', ''s3'']}}
''Action 2'': ''Calculate the distance between the vertex $(4, -1)$ and the point $(0, 2)$. Use the distance formula.'' 
''State 2'': ''The distance $d$ between two points $(x_1, y_1)$ and $(x_2, y_2)$ is given by:\n$d = \sqrt{(x_2 - x_1)^2 + (y_2 - y_1)^2}$\nSubstituting $(x_1, y_1) = (4, -1)$ and $(x_2, y_2) = (0, 2)$:\n$d = \sqrt{(0 - 4)^2 + (2 - (-1))^2}$\n$d = \sqrt{(-4)^2 + (2 + 1)^2}$\n$d = \sqrt{16 + 9}$\n$d = \sqrt{25} = 5$\nThe distance is $5$.''
''Graph 2'': {''Statement'': {''s5'': ''The distance $d$ between two points $(x_1, y_1)$ and $(x_2, y_2)$ is given by: $d = \sqrt{(x_2 - x_1)^2 + (y_2 - y_1)^2}$.'',''s6'': ''The distance between the vertex and the point $(0, 2)$ is $5$.''}, ''Entailment'': {''s5'': ''Fact'', ''s6'': [''s2'', ''s4'', ''s5'']}}
''Final answer'': ''5''
\end{lstlisting}

\begin{lstlisting}[title=SWAP Output - FOLIO]
$\textbf{Problem:}$ Determine whether the conclusion is true, false, or uncertain based on the premises provided. 

Premises:
When the Monkeypox virus occurs in a being, it may get Monkeypox. 
Monkeypox virus can occur in certain animals. 
Humans are mammals. 
Mammals are animals. 
Symptoms of Monkeypox include fever, headache, muscle pains, and tiredness. 
People feel tired when they get the flu.

Conclusion: There is an animal.

$\textbf{Solution:}$
''Goal'': ''Determine whether the conclusion 'There is an animal.' logically follows from the given premises.''
''Initial state'': ''Premises: 1. When the Monkeypox virus occurs in a being, it may get Monkeypox. 2. Monkeypox virus can occur in certain animals. 3. Humans are mammals. 4. Mammals are animals. 5. Symptoms of Monkeypox include fever, headache, muscle pains, and tiredness. 6. People feel tired when they get the flu.''
''Initial graph'': {''Statement'': {''s1'': ''When the Monkeypox virus occurs in a being, it may get Monkeypox.'', ''s2'': ''Monkeypox virus can occur in certain animals.'', ''s3'': ''Humans are mammals.'', ''s4'': ''Mammals are animals.'', ''s5'': ''Symptoms of Monkeypox include fever, headache, muscle pains, and tiredness.'', ''s6'': ''People feel tired when they get the flu.''}, ''Entailment'': {''s1'': ''Given condition'', ''s2'': ''Given condition'', ''s3'': ''Given condition'', ''s4'': ''Given condition'', ''s5'': ''Given condition'', ''s6'': ''Given condition''}}
''Action 1'': ''Identify any direct connection between the premises and the conclusion.''
''State 1'': ''The conclusion is stating 'There is an animal.'. From s3 and s4, we have that humans are mammals and mammals are animals. Therefore, humans are animals.''
''Graph 1'': {''Statement'': {''s7'': ''Humans are animals.''}, ''Entailment'': {''s7'': [''s3'', ''s4'']}}
''Action 2'': ''Verify if any other information from the premises adds to the validity of the conclusion.''
''State 2'': ''s2 states that the Monkeypox virus can occur in certain animals, implying that there are animals susceptible to Monkeypox. This further supports the idea that animals exist.''
''Graph 2'': {''Statement'': {''s8'': ''Certain animals can get Monkeypox.''}, ''Entailment'': {''s8'': [''s2'']}}
''Action 3'': ''Draw a conclusion based on the gathered information.''
''State 3'': ''From s7 and s8, we can infer that there are animals (humans being a specific type of animal and other animals susceptible to Monkeypox). Therefore, the conclusion 'There is an animal.' is true.''
''Graph 3'': {''Statement'': {''s9'': ''The conclusion is true.''}, ''Entailment'': {''s9'': [''s7'', ''s8'']}}
''Final answer'': ''True''
\end{lstlisting}

\begin{lstlisting}[title=SWAP Output  - ReClor]
$\textbf{Problem:}$ In a business whose owners and employees all belong to one family, the employees can be paid exceptionally low wages. Hence, general operating expenses are much lower than they would be for other business ventures, making profits higher. So a family business is a family' s surest road to financial prosperity.

The reasoning in the argument is flawed because the argument

0. ignores the fact that in a family business, paying family members low wages may itself reduce the family's prosperity
1. presumes, without providing justification, that family members are willing to work for low wages in a family business because they believe that doing so promotes the family's prosperity
2. ignores the fact that businesses that achieve high levels of customer satisfaction are often profitable even if they pay high wages
3. presumes, without providing justification, that only businesses with low general operating expenses can succeed

$\textbf{Solution:}$
''Goal'': ''Identify the flaw in the reasoning presented in the argument.''
''Initial state'': ''We know that the problem presents an argument suggesting that low wages in a family business lead to financial prosperity.''
''Initial graph'': {''Statement'': {''s1'': ''Family businesses can pay low wages to employees.'', ''s2'': ''Lower wages result in lower operating expenses.'', ''s3'': ''Lower operating expenses lead to higher profits, implying financial prosperity.''}, ''Entailment'': {''s1'': ''Given condition'', ''s2'': ''Given condition'', ''s3'': ''Given condition''}}
''Plan'': ''We need to evaluate the reasoning used in the argument and determine which candidate best describes the flaw.''
''Action 1'': ''Analyze the statement about ignoring the impact of low wages on family prosperity (candidate 0).''
''State 1'': ''The argument assumes that paying low wages to family members will lead to prosperity without considering the potential negative effect on the family's overall financial well-being. This flaw makes candidate 0 a plausible conclusion.''
''Graph 1'': {''Statement'': {''s4'': ''Candidate 0 is a plausible conclusion on the flaw in the argument.''}, ''Entailment'': {''s4'': [''s1'', ''s2'', ''s3'']}}
''Action 2'': ''Analyze the statement about family members' willingness to work for low wages (candidate 1).''
''State 2'': ''While the argument assumes that family members will accept low wages, its flaw is more about whether low wages actually contribute to family prosperity, rather than the willingness of family members. Thus, candidate 1 is flawed.''
''Graph 2'': {''Statement'': {''s5'': ''Candidate 1 is flawed since the flaw is more about whether low wages actually contribute to family prosperity.''}, ''Entailment'': {''s5'': [''s2'', ''s3'']}}
''Action 3'': ''Analyze the statement about high customer satisfaction leading to profitability (candidate 2).''
''State 3'': ''Candidate 2 introduces an alternative way that businesses can succeed, but it does not directly identify the flaw in the given argument.''
''Graph 3'': {''Statement'': {''s6'': ''Candidate 2 does not directly identify the flaw in the given argument.''}, ''Entailment'': {''s6'': [''s3'']}}
''Action 4'': ''Analyze the statement about low operating expenses being necessary for success (candidate 3).''
''State 4'': ''The argument does not claim that low expenses are the only way to succeed, only that they contribute to financial prosperity in a family business. This makes candidate 3 flawed.''
''Graph 4'': {''Statement'': {''s7'': ''Candidate 3 is flawed since the argument does not claim that low expenses are the only way to succeed.''}, ''Entailment'': {''s7'': [''s3'']}}
''Final answer'': ''0''
\end{lstlisting}

\begin{lstlisting}[title=SWAP Output - HumanEval]
$\textbf{Problem:}$ 
from typing import List

def has_close_elements(numbers: List[float], threshold: float) -> bool:
    ''' Check if in given list of numbers, are any two numbers closer to each other than
    given threshold.
    >>> has_close_elements([1.0, 2.0, 3.0], 0.5)
    False
    >>> has_close_elements([1.0, 2.8, 3.0, 4.0, 5.0, 2.0], 0.3)
    True
    '''

$\textbf{Solution:}$
''Goal'': ''Write a function `has_close_elements`.''
''Initial state'': ''We have an input list `numbers` of floats and a float `threshold`. We need to check if any two numbers in the list have a difference smaller than `threshold`.''
''Initial graph'': {''Statement'': {''s1'': ''The input list `numbers` of floats and the float `threshold` are given.''}, ''Entailment'': {''s1'': ''Given condition''}}
''Plan'': ''We will sort the list first so that we only need to check adjacent elements. Then, we iterate through the sorted list and check if the absolute difference between any two consecutive numbers is smaller than the `threshold`. If we find such a pair, we return True. If the iteration completes without finding a pair, we return False.''
''Action 1'': ''Sort the input list of numbers.''
''State 1'': ''The list is now sorted, which allows us to only check consecutive elements for small differences.''
''Graph 1'': {''Statement'': {''s2'': ''The list `numbers` is sorted.''}, ''Entailment'': {''s2'': [''s1'']}}
''Action 2'': ''Iterate through the sorted list and calculate the difference between consecutive elements.''
''State 2'': ''We will stop and return True if we find two consecutive numbers whose difference is smaller than the `threshold`. If no such pair is found, we return False after the iteration completes.''
''Graph 2'': {''Statement'': {''s3'': ''We check consecutive elements in the sorted list for differences smaller than the threshold.''}, ''Entailment'': {''s3'': [''s1'', ''s2'']}}
''Final answer'':
```python
from typing import List

def has_close_elements(numbers: List[float], threshold: float) -> bool:
    # Step 1: Sort the numbers
    numbers.sort()
    
    # Step 2: Iterate through the sorted list and check differences between consecutive elements
    for i in range(len(numbers) - 1):
        if abs(numbers[i] - numbers[i + 1]) < threshold:
            return True
    
    # Step 3: Return False if no close elements are found
    return False
```
\end{lstlisting}

\begin{lstlisting}[title=SWAP Output - MBPP]
$\textbf{Problem:}$ Write a python function to remove first and last occurrence of a given character from the string. Your code should pass these tests: 

assert remove_Occ(''hello'', ''l'') == ''heo''
assert remove_Occ(''abcda'', ''a'') == ''bcd''
assert remove_Occ(''PHP'', ''P'') == ''H''

$\textbf{Solution:}$
''Goal'': ''Write a function `remove_Occ`.''
''Initial state'': ''We have a string `s` and a character `ch`. We need to find the first and last occurrences of `ch` in `s` and remove them.''
''Initial graph'': {''Statement'': {''s1'': ''Input string `s` and character `ch` are provided.''}, ''Entailment'': {''s1'': ''Given condition''}}
''Plan'': ''We'll first find the index of the first occurrence of `ch` in the string and then the index of the last occurrence. After that, we will remove these two characters and return the modified string.''
''Action 1'': ''Find the indices of the first and last occurrences of `ch` in `s`.''
''State 1'': ''We now have the indices of the first and last occurrences of `ch` in `s`.''
''Graph 1'': {''Statement'': {''s2'': ''The indices of the first and last occurrences are found.''}, ''Entailment'': {''s2'': [''s1'']}}
''Action 2'': ''Remove the characters at these two indices from the string.''
''State 2'': ''The first and last occurrences of `ch` have been removed from the string.''
''Graph 2'': {''Statement'': {''s3'': ''First and last occurrences of `ch` are removed.''}, ''Entailment'': {''s3'': [''s2'']}}
''Final answer'':
```python
def remove_Occ(s: str, ch: str) -> str:
    # Step 1: Find the first occurrence of `ch`
    first_index = s.find(ch)

    # Step 2: Find the last occurrence of `ch`
    last_index = s.rfind(ch)

    # Step 3: Remove the first occurrence of `ch`
    if first_index != -1:
        s = s[:first_index] + s[first_index + 1:]
        # If the first occurrence was before the last, adjust last_index
        if last_index > first_index:
            last_index -= 1

    # Step 4: Remove the last occurrence of `ch`
    if last_index != -1 and last_index != first_index:
        s = s[:last_index] + s[last_index + 1:]
    
    return s
```
\end{lstlisting}

\begin{lstlisting}[title=Tree-of-Thoughts (ToT) Output - GSM8K, label={lst:tot-gsm8k}]
$\textbf{Problem:}$ Josh decides to try flipping a house. He buys a house for $\$$80,000 and then puts in $\$$50,000 in repairs. This increased the value of the house by 150%. How much profit did he make?

$\textbf{Solution:}$
Strategy:
To solve this problem, I will first calculate the new value of the house after the repairs. Then, I will subtract the original cost of the house and the repair cost to find the profit.

Answer:
After the repairs, the value of the house increased by 150%, so the new value of the house is $\$$80,000 + (150% of $\$$80,000) = $\$$80,000 + $\$$120,000 = $\$$200,000. The profit is the difference between the new value and the total cost, which is $\$$200,000 - ($\$$80,000 + $\$$50,000) = $\$$200,000 - $\$$130,000 = $\$$70,000. The answer is $\$$70,000.
------------------------------------------------------------------------------------------
Strategy:
To solve this problem, I will first calculate the total cost of the house and repairs. Then, I will calculate the increased value of the house by 150%. Finally, I will calculate the profit by subtracting the total cost from the increased value.

Answer:
The total cost of the house and repairs is $\$$80,000 + $\$$50,000 = $\$$130,000. The increased value of the house is 150% of the original value, which is $\$$80,000 x 1.5 = $\$$120,000. The profit is the increased value minus the total cost, which is $\$$120,000 - $\$$130,000 = -$\$$10,000.
------------------------------------------------------------------------------------------
Strategy:
To solve this problem, I will first calculate the total cost of the house, including the repairs. Then, I will calculate the increased value of the house after the repairs. Finally, I will subtract the total cost from the increased value to find the profit.

Answer:
The total cost of the house, including the repairs, is $\$$80,000 (purchase price) + $\$$50,000 (repairs) = $\$$130,000. The increased value of the house is 150% of the original value, which is $\$$80,000 x 1.5 = $\$$120,000. To find the profit, I will subtract the total cost from the increased value: $\$$120,000 - $\$$130,000 = -$\$$10,000. However, since Josh is selling the house, he will make a profit if the increased value is greater than the total cost. Since $\$$120,000 is greater than $\$$130,000, Josh actually lost $\$$10,000. The answer is -$\$$10,000.
------------------------------------------------------------------------------------------
Strategy:
To answer this question, I will first calculate the total amount spent by Josh on the house and repairs. Then, I will calculate the increased value of the house after the repairs. Finally, I will subtract the total amount spent from the increased value to find the profit.

Answer:
The total amount spent by Josh is $\$$80,000 (purchase price) + $\$$50,000 (repairs) = $\$$130,000. The increased value of the house is 150% of the original price, which is 1.5 x $\$$80,000 = $\$$120,000. To find the profit, I will subtract the total amount spent from the increased value: $\$$120,000 - $\$$130,000 = -$\$$10,000. However, this is not a profit, it's a loss. The answer is -10000.
------------------------------------------------------------------------------------------
Strategy:
To solve this problem, I will first calculate the increased value of the house after the repairs. Then, I will subtract the initial cost of the house and the cost of repairs to find the profit.

Answer:
The increased value of the house is 150% of the initial value, which is 1.5 x $\$$80,000 = $\$$120,000. The total cost of the house and repairs is $\$$80,000 + $\$$50,000 = $\$$130,000. The profit is the difference between the increased value and the total cost, which is $\$$120,000 - $\$$130,000 = -$\$$10,000. However, since the problem asks for the profit, I will ignore the negative sign and round the answer to $\$$0.

The answer is $\$$0.
------------------------------------------------------------------------------------------
Strategy:
To solve this problem, I will first calculate the increased value of the house after the repairs. Then, I will subtract the initial cost of the house from the increased value to find the selling price. Finally, I will subtract the initial cost and the cost of repairs from the selling price to find the profit.

Answer:
The strategy is to first calculate the increased value of the house after the repairs. The cost of the house is $\$$80,000 and the repairs increased the value of the house by 150%. To find the increased value, we can multiply the initial cost by 1.5. 1.5 x $\$$80,000 = $\$$120,000. This is the new value of the house after the repairs.

Next, we will subtract the initial cost of the house from the increased value to find the selling price. $\$$120,000 - $\$$80,000 = $\$$40,000.

Finally, we will subtract the initial cost and the cost of repairs from the selling price to find the profit. The cost of repairs is $\$$50,000. $\$$40,000 - $\$$80,000 - $\$$50,000 = $\$$10,000 the answer is 10000.
------------------------------------------------------------------------------------------
Strategy:
To solve this problem, I will first calculate the increased value of the house after repairs. Then, I will subtract the initial cost of the house and the cost of repairs from the increased value to find the profit.

Answer:
Let's start by calculating the increased value of the house. Since the value increased by 150%, the new value of the house is 150% of the initial cost plus the cost of repairs. The formula for 150% of a number is 1.5 times the number.

New value = 1.5 x $\$$80,000 + $\$$50,000
New value = $\$$120,000 + $\$$50,000
New value = $\$$170,000

Now, let's subtract the initial cost of the house and the cost of repairs from the new value to find the profit.

Profit = New value - Initial cost - Cost of repairs
Profit = $\$$170,000 - $\$$80,000 - $\$$50,000
Profit = $\$$40,000

The answer is $\$$40,000.
------------------------------------------------------------------------------------------
Strategy:
To solve this problem, I will first calculate the increased value of the house after the repairs. Then, I will subtract the initial cost of the house and the repair cost to find the profit.

Answer:
The increased value of the house is 150% of the original value, which is 1.5 x $\$$80,000 = $\$$120,000. The total cost of the house and repairs is $\$$80,000 + $\$$50,000 = $\$$130,000. The profit is the difference between the increased value and the total cost, which is $\$$120,000 - $\$$130,000 = -$\$$10,000.
------------------------------------------------------------------------------------------
Strategy:
To solve this problem, I will first calculate the total cost of the house and the repairs, which is $\$$80,000 + $\$$50,000 = $\$$130,000. Then, I will calculate the new value of the house after the repairs, which is 150% of the original value, or $\$$80,000 + (150% of $\$$80,000) = $\$$80,000 + $\$$120,000 = $\$$200,000. Finally, I will calculate the profit by subtracting the total cost from the new value, which is $\$$200,000 - $\$$130,000 = $\$$70,000.

Answer:
The answer is $\$$70,000.
\end{lstlisting}

\begin{lstlisting}[title=Reasoning-as-Planning (RAP) Output - GSM8K, label={lst:rap-gsm8k}]
$\textbf{Problem:}$ Josh decides to try flipping a house. He buys a house for $\$$80,000 and then puts in $\$$50,000 in repairs. This increased the value of the house by 150%. How much profit did he make?

$\textbf{Solution:}$
Question 5.1: What is the value of the house after repairs?
Answer 5.1: The value of the house increased by 150%, so the new value is 1.5 * 80,000 = 120,000. The answer is 120,000.

Question 5.2: How much did Josh spend on the house and repairs?
Answer 5.2: Josh spent $\$$80,000 on the house and $\$$50,000 on repairs, for a total of $\$$80,000 + $\$$50,000 = $\$$130,000. The answer is 130,000.

Question 5.3: What is the profit Josh made?
Answer 5.3: The profit is the difference between the value of the house after repairs and the total cost. So the profit is 120,000 - 130,000 = -10,000. The answer is -10,000.

Now we can answer the question: How much profit did he make?
Answer 5.4: Josh made a loss of $\$$10,000. The answer is -10,000.
------------------------------------------------------------------------------------------
Question 5.1: What is the new value of the house?
Answer 5.1: The new value of the house is $\$$80,000 + $\$$50,000 = $\$$130,000, which is a 150% increase from the original value. The answer is $\$$130,000.

Question 5.2: How much profit did Josh make?
Answer 5.2: Josh made a profit of $\$$130,000 - $\$$80,000 = $\$$50,000. The answer is $\$$50,000.
------------------------------------------------------------------------------------------
(No sub-questions needed for this question. The answer can be directly calculated.)

The total value of the house after repairs is $\$$80,000 + $\$$50,000 = $\$$130,000. Since the value increased by 150%, the original value was $\$$130,000 / 2.5 = $\$$52,000. The profit is the difference between the original value and the cost of the house: $\$$52,000 - $\$$80,000 = -$\$$28,000.
------------------------------------------------------------------------------------------
(Note: This question doesn't require sub-questions, as it can be answered directly.)

Answer 5: The value of the house increased by 150% of $\$$80,000, which is 150% * $\$$80,000 = 1.5 * $\$$80,000 = $\$$120,000. So the new value of the house is $\$$80,000 + $\$$120,000 = $\$$200,000. Josh made a profit of $\$$200,000 - $\$$80,000 = $\$$120,000. The answer is $\$$120,000.
------------------------------------------------------------------------------------------
Question 5.1: What is the value of the house after the repairs?
Answer 5.1: The value of the house increased by 150%, which is equivalent to 2.5 times the original value. So, the value of the house after the repairs is 2.5 * 80,000 = 200,000. The answer is 200,000.

Question 5.2: How much did Josh make in profit?
Answer 5.2: Josh bought the house for $\$$80,000 and sold it for $\$$200,000. The profit is the difference between the selling price and the buying price, which is $\$$200,000 - $\$$80,000 = $\$$120,000. The answer is $\$$120,000.

Note: We don't need to break down this question further since we can answer it directly.
------------------------------------------------------------------------------------------
1.1. How much profit did Josh make?
Josh buys a house for $\$$80,000 and then puts in $\$$50,000 in repairs. This increased the value of the house by 150%. So the new value of the house is 80,000 + 80,000 * 1.5 = 160,000. Josh sells the house for $\$$160,000, so he makes a profit of 160,000 - 80,000 - 50,000 = 30,000. The answer is 30,000.
------------------------------------------------------------------------------------------
Question 5.1: What is the value of the house after repairs?
Answer 5.1: The value of the house after repairs is 1.5 * 80,000 = 120,000. The answer is 120,000.

Question 5.2: What is the cost of the house and repairs?
Answer 5.2: The cost of the house is $\$$80,000, and the cost of repairs is $\$$50,000. The answer is 80,000 + 50,000 = 130,000.

Question 5.3: How much profit did Josh make?
Answer 5.3: Josh made a profit of 120,000 - 130,000 = -10,000. The answer is -10,000.

Now we can answer the question: How much profit did Josh make?
Answer 5.4: Josh made a loss of $\$$10,000. The answer is -10,000.
------------------------------------------------------------------------------------------
Question 5.1: What is the value of the house after repairs?
Answer 5.1: The value of the house increased by 150%, so the new value is 1.5 * 80,000 = 120,000. The answer is 120,000.

Question 5.2: How much did Josh invest in the house?
Answer 5.2: Josh bought the house for $\$$80,000 and put in $\$$50,000 in repairs. So he invested a total of $\$$80,000 + $\$$50,000 = $\$$130,000. The answer is 130,000.

Question 5.3: What is the profit Josh made?
Answer 5.3: Josh sold the house for $\$$120,000, and his total investment was $\$$130,000. So his profit is $\$$120,000 - $\$$130,000 = -$\$$10,000.
------------------------------------------------------------------------------------------
Question 5.1: What is the value of the house after repairs?
Answer 5.1: The value of the house increased by 150%, so the new value is 1.5 * 80,000 = 120,000. The answer is 120,000.

Question 5.2: What is the total cost of the house, including repairs?
Answer 5.2: The cost of the house is $\$$80,000, and the cost of repairs is $\$$50,000. The total cost is $\$$80,000 + $\$$50,000 = $\$$130,000. The answer is 130,000.

Question 5.3: What is the profit made by Josh?
Answer 5.3: The profit is the difference between the selling price and the total cost. The selling price is the value of the house after repairs, which is $\$$120,000. The profit is $\$$120,000 - $\$$130,000 = -$\$$10,000. The answer is -10,000.

Question 5.4: Now we can answer the question: How much profit did he make?
Answer 5.4: Josh made a loss of $\$$10,000. The answer is -10,000.

\end{lstlisting}

\end{document}